\documentclass{tlp}

\usepackage{graphicx}
\usepackage{multirow}
\usepackage{times}  
\usepackage{helvet}  
\usepackage{courier}  

\usepackage{amsmath}
\usepackage{amssymb}
\usepackage{amsfonts}
\usepackage{latexsym}
\usepackage{url}
\usepackage{array}
\usepackage{bm}
\usepackage{wrapfig}
\newfont{\Bb}{msbm10}
\usepackage[dvipdfmx]{}
\newcommand{\ignore}[1]{}
\renewcommand{\phi}{\varphi}
\renewcommand{\emptyset}{\varnothing}
\newcommand{\pf}[1]{\langle\,#1\,\rangle}

\newtheorem{proposition}{Proposition}[section]

\newtheorem{example}{Example}[section]
\newtheorem{definition}{Definition}[section]

\begin{document}

\lefttitle{Cambridge Author}

\jnlPage{1}{34}
\jnlDoiYr{2024}
\doival{10.1017/S1471068423000376}

\title[Human Conditional Reasoning in Answer Set Programming]
{Human Conditional Reasoning in Answer Set Programming}

\begin{authgrp}
\author{\sn{Chiaki Sakama}}
\affiliation{Wakayama University, 930 Sakaedani, Wakayama 640-8510, Japan\\
sakama@wakayama-u.ac.jp}
\end{authgrp}

\history{{\rm This paper has been published in} Theory and Practice of Logic Programming, 24(1), January 2024, pp.\ 157--192.}

\maketitle

\begin{abstract}
Given a conditional sentence ``$\phi\Rightarrow \psi$" (if $\phi$ then $\psi$) 
and respective facts, four different types of inferences are observed in human reasoning. 
{\em Affirming the antecedent\/} (AA) (or {\em modus ponens\/}) reasons $\psi$ from $\phi$; 
{\em affirming the consequent\/} (AC) reasons $\phi$ from $\psi$; 
{\em denying the antecedent\/} (DA) reasons $\neg\psi$ from $\neg\phi$; and 
{\em denying the consequent\/} (DC) (or {\em modus tollens\/}) reasons $\neg\phi$ from $\neg\psi$. 
Among them, AA and DC are logically valid, while AC and DA are logically invalid and often called {\em logical fallacies}. 
Nevertheless, humans often perform AC or DA as {\em pragmatic inference\/} in daily life. 
In this paper, we realize AC, DA and DC inferences in {\em answer set programming}. 
Eight different types of {\em completion\/} are introduced and their semantics are given by answer sets. 
We investigate formal properties and characterize human reasoning tasks in cognitive psychology.  
Those completions are also applied to commonsense reasoning in AI. 
\end{abstract}

\begin{keywords}
answer set programming, completion, human conditional reasoning, pragmatic inference
\end{keywords}

\section{Introduction} \label{sec:1}
People use conditional sentences and reason with them in everyday life. 
From an early stage of artificial intelligence (AI), researchers represent conditional sentences as if-then rules and perform deductive inference using them. 
Production systems or logic programming are examples of this type of systems. 
However, human conditional reasoning is not always logically valid.  
In psychology and cognitive science, it is well known that humans are more likely to perform logically invalid but pragmatic inference. 
For instance, consider the following three sentences:  
\begin{description}\setlength{\itemsep}{-0.3mm}
\item[]$S$:\,  If the team wins the first round tournament, then it advances to the final round. 
\item[]$P$:\,  The team wins the first round tournament.
\item[]$C$:\, The team advances to the final round. 
\end{description}
Given the conditional sentence $S$ and the premise $P$, {\em affirming the antecedent\/} (AA) (or {\em modus ponens\/}) 
concludes the consequence $C$.  Given $S$ and the negation of the consequence $\neg\, C$, 
{\em denying the consequent\/} (DC) (or {\em modus tollens\/}) concludes the negation of the premise $\neg\, P$. 
AA and DC are logically valid. 
On the other hand,  people often infer $P$ from $S$ and $C$ or infer $\neg\, C$ from $S$ and $\neg\, P$. 
The former is called {\em affirming the consequent\/} (AC) and the latter is called 
{\em denying the antecedent\/} (DA).  Both AC and DA are logically invalid and often called {\em logical fallacies}. 

In the pragmatics of conditional reasoning, it is assumed that a conditional sentence is often interpreted as bi-conditional, that is, 
 `{\em if\/}' is interpreted as  `{\em if and only if\/}', and such {\em conditional perfection\/} 
produces AC or DA as {\em invited inference\/} (\cite{GZ71,Horn00}).  
Psychological studies empirically show that a conditional sentence ``$p$ {\em if\/} $q$" is rephrased into the form ``$p$ {\em only if\/} $q$" with 
greater frequency for permission/obligation statements (\cite{CH85,Byr05}). 
For instance, the sentence 
``a customer can drink an alcoholic beverage if he is over 18" is rephrased into 
``a customer can drink an alcoholic beverage {\em only if\/} he is over 18". 
It is also reported that AA is easier than DC when a conditional is given as ``{\em if\/} $p$ {\em then\/} $q$". 
When a conditional is given as ``$p$ {\em only if\/} $q$", on the other hand, it is rephrased as ``{\em if not\/} $q$ 
{\em then not\/} $p$" and this paraphrase yields a directionality opposite which makes DC easier than AA (\cite{Bra78}). 
The fact that people do not necessarily make inferences as in standard logic 
brings several proposals of new interpretation of conditional sentences in cognitive psychology. 
{\em Mental logic\/} (\cite{BO98}) interprets `{\em if\/}' as conveying supposition and 
introduces a set of pragmatic inference schemas for {\em if}-conditionals.  
{\em Mental model theory\/} (\cite{JL83}), on the other hand, considers that the meanings of conditionals are not truth-functional, and 
represents the meaning of a conditional sentence by models of the possibilities compatible with the sentence. 
A probabilistic approach interprets a conditional sentence ``$p\Rightarrow q$" in terms of conditional probability $P(q\mid p)$, then  
the acceptance rates of four conditional inferences are represented by their respective conditional probabilities (\cite{OC01}). 
\cite{EKR18} use {\em conditional logic\/} and define {\em inference patterns\/} 
as combination of four inference rules (AA, DC, AC, DA). 
Given a conditional sentence ``{\em if\/} $p$ {\em then\/} $q$", four possible worlds (combination of truth values of $p$ and $q$) are 
considered. 
An inference in each pattern is then defined by imposing corresponding constraints on the plausibility relation over the worlds. 

In this way, the need of considering the pragmatics of conditional reasoning has been widely recognized in psychology and cognitive science. On the other hand, relatively little attention has been paid for realizing such pragmatic inference in computational logic or logic programming (\cite{SvL08,Kow11}). 
From a practical perspective, however, people would expect AI to reason like humans, that is, one would expect AI to conclude 
$P$ from $S$ and $C$, or $\neg\, C$ from $S$ and $\neg\, P$ in the introductory example, rather than conclude $unknown$. 
Logic programming is a context-independent language and has a general-purpose inference mechanism by its nature. 
By contrast, pragmatic inference is governed by context-sensitive mechanisms, 
rather than context-free and general-purpose mechanisms (\cite{CH85,CT92}).  
As argued by \cite{DHR12}, 
computational approaches to explain human reasoning should be {\em cognitively adequate}, that is, they appropriately 
represent human knowledge ({\em conceptually adequate\/}) and computations behave similarly to human reasoning 
({\em inferentially adequate\/}). 
Then if we use logic programming for representing knowledge in daily life, it is useful to have a mechanism of automatic 
transformation of a knowledge base to simulate human reasoning depending on the context in which conditional sentences are used. 
That is, transform a program to a conceptually adequate form in order to make computation in the program inferentially adequate. 

In this paper, we realize human conditional reasoning in {\em answer set programming\/} (ASP) (\cite{GL91}). 
ASP is one of the most popular frameworks that realize declarative knowledge representation and commonsense reasoning. 
ASP is a language of logic programming and conditional sentences are represented by rules in a program. 
Inference in ASP is deduction based on {\em default logic\/} (\cite{Rei80}), while modus tollens or DC is not considered in ASP. 
AC and DA are partly realized by {\em abductive logic programming\/} (\cite{KKT92}) and 
{\em program completion\/} (\cite{Cla78}), respectively.  
As will be argued in this paper, however, AC and DA produce different results from them in general. 
We realize pragmatic AC and DA inferences as well as DC inference in ASP in a uniform and modular way. 
We introduce the notions of {\em AC completion\/}, {\em DC completion\/}, {\em DA completion\/} and their variants. 
We investigate formal properties of those completions and characterize human reasoning tasks in cognitive psychology. 
We also address applications to commonsense reasoning in AI. 
The rest of this paper is organized as follows. 
Section~2 reviews basic notions of ASP programs considered in this paper. 
Section~3 introduces different types of completions for human conditional reasoning, 
and Section~4 presents their variants as default reasoning. 
Section~5 characterizes human reasoning tasks in the literature, and Section~6 
addresses applications to commonsense reasoning. 
Section~7 discusses related works and Section~8 summarizes the paper. 

\section{Preliminaries}  \label{sec:2}

In this paper, we consider logic programs with disjunction, default negation, and explicit negation. 
A {\em general extended disjunctive program\/} (GEDP) (\cite{LW92,IS98}) $\Pi$ is a set of rules of the form: 
\begin{eqnarray}\label{pnp}
&& L_1\,;\,\cdots\,;\, L_k\, ;\, not\,L_{k+1}\,;\cdots ;\,not\,L_l \nonumber\\
&& \quad\qquad\leftarrow\; L_{l+1},\,\ldots,\,L_m,\,not\,L_{m+1},\,\ldots,\,not\,L_n\;\;
\end{eqnarray}
where $L_i$'s $(1\le i\le n)$ 
are (positive or negative) literals and $0\leq k\leq l\leq m\leq n$. 
A program might contain two types of negation: default negation (or negation as failure) $not$ and explicit negation $\neg$.  
For any literal $L$, $not\,L$ is called an {\em NAF-literal\/} and define $\neg\neg L=L$. 
We often use the letter $\ell$ to mean either a literal $L$ or an NAF-literal $not\,L$. 
The left of ``$\leftarrow$" is a disjunction of literals and NAF-literals (called {\em head\/}), and 
the right of ``$\leftarrow$" is a conjunction of literals and NAF-literals (called {\em body\/}).  
Given a rule $r$ of the form~(\ref{pnp}), define  
$head^+(r)=\{L_1,\ldots,L_k\}$, $head^-(r)=\{L_{k+1},\ldots,L_l\}$, 
$body^+(r)=\{L_{l+1},\ldots,L_m\}$, and $body^-(r)=\{L_{m+1},\ldots,L_n\}$. 
A rule~(\ref{pnp}) is called a {\em fact\/} if $body^+(r)=body^-(r)=\emptyset$; and it is called a {\em constraint\/} if 
$head^+(r)=head^-(r)=\emptyset$.  
A GEDP $\Pi$ is called $not$-{\em free\/} if $head^-(r)=body^-(r)=\emptyset$ for each rule $r$ in $\Pi$. 

A GEDP $\Pi$ coincides with an {\em extended disjunctive program\/} (EDP) of (\cite{GL91}) 
if $head^-(r)=\emptyset$ for any rule $r$ in $\Pi$. 
An EDP $\Pi$ is called (i) an {\em extended logic program\/} (ELP) if $\mid\!\! head^+(r)\!\!\mid\,\le 1$ for any $r\in\Pi$; 
and (ii) a {\em normal disjunctive program\/} (NDP) if $\Pi$ contains no negative literal. 
An NDP $\Pi$ is called (i) a {\em positive disjunctive program\/} (PDP) if $\Pi$ contains no NAF-literal; 
and (ii) a {\em normal logic program\/} (NLP) if $\mid\!\! head^+(r)\!\!\mid\,\le 1$ for any $r\in\Pi$. 
In this paper, we consider ground programs containing no variable and 
a {\em program\/} means a (ground) GEDP unless stated otherwise. 

Let $Lit$ be the set of all ground literals in the language of a program.  
A set of ground literals $S\subseteq Lit$ {\em satisfies\/} a ground rule $r$ of the form~(\ref{pnp}) iff 
$body^+(r)\subseteq S$ and $body^-(r)\cap S=\emptyset$ imply either 
$head^+(r)\cap S\neq\emptyset$ or $head^-(r)\not\subseteq S$. 
In particular, when $head^+(r)=head^-(r)=\emptyset$, $S\subseteq Lit$ satisfies a constraint $r$ iff 
$body^+(r)\not\subseteq S$ or $body^-(r)\cap S\neq\emptyset$. 
The {\em answer sets\/} of a GEDP are defined by the following two steps.  
First, let $\Pi$ be a $not$-free GEDP and $S\subseteq Lit$.  
Then, $S$ is an {\em answer set\/} of $\Pi$ 
iff $S$ is a minimal set satisfying the conditions: 
(i) $S$ satisfies every rule from $\Pi$, that is, for each ground rule:
\begin{equation}\label{not-free}
L_1\, ;\,\cdots\, ;\, L_k\leftarrow L_{l+1},\,\ldots,\,L_m
\end{equation}
 from $\Pi$, $\{L_{l+1},\ldots,L_m\}\subseteq S$ implies $\{L_1,\ldots,L_k\}\cap S\ne\emptyset$. 
(ii) If $S$ contains a pair of complementary literals $L$ and $\neg L$, then $S=Lit$.%
\footnote{By this definition, an answer set is not {\em paraconsistent\/}, i.e., $\{L,\neg L\}\subseteq S$ makes $S$ a trivial set $Lit$. 
A paraconsistent semantics of EDPs is given in (\cite{SI95}).}

Second, let $\Pi$ be any GEDP and $S\subseteq Lit$. 
The {\em reduct\/} $\Pi^S$ of $\Pi$ by $S$ is 
a $not$-free EDP obtained as follows: a rule $r^S$ of the form~(\ref{not-free}) 
is in $\Pi^S$ iff there is a ground rule $r$ of the form~(\ref{pnp}) from $\Pi$ such that 
$head^-(r)\subseteq S$ and $body^-(r)\cap S=\emptyset$. 
For programs of the form $\Pi^S$, their answer sets have already been defined. 
Then, $S$ is an {\em answer set\/} of $\Pi$ iff $S$ is an answer set of $\Pi^S$. 

When a program $\Pi$ is an EDP, the above definition of answer sets coincides with that given in (\cite{GL91}).  
It is shown that every answer set of a GEDP $\Pi$ satisfies every rule from $\Pi$ (\cite{IS98}).  
An answer set is {\em consistent\/} if it is not $Lit$.  
A program $\Pi$ is {\em consistent\/} if it has a consistent answer set; otherwise, $\Pi$ is {\em inconsistent}. 
When a program $\Pi$ is inconsistent, there are two different cases.  If $\Pi$ has the single answer set $Lit$, $\Pi$ is called 
 {\em contradictory\/}; else if $\Pi$ has no answer set, $\Pi$ is called {\em incoherent\/}.  
 The difference of two cases is illustrated by the following example. 
 
 \begin{example}\rm 
The program $\Pi_1=\{\,p\leftarrow not\,q,\;\; \neg p\leftarrow \,\}$ is incoherent, while 
$\Pi_2=\{\,p\leftarrow q,\;\;  q\leftarrow,\;\; \neg p\leftarrow\,\}$ is contradictory. 
Note that $Lit$ is not the answer set of $\Pi_1$ because $Lit$ is not the answer set of $\Pi_1^{Lit}=\{\,\neg p\leftarrow\,\}$. 
\end{example} 
 
We write $\Pi\models_c L$ (resp.\ $\Pi\models_s L$ ) if a literal $L$ is included in some (resp.\ every) consistent answer set of $\Pi$.%
\footnote{$\models_c$ (resp.\ $\models_s$) means entailment under credulous (resp.\ skeptical) reasoning.}
Two programs $\Pi_1$ and $\Pi_2$ are {\em equivalent\/} if they have the same set of answer sets. 
Two programs $\Pi_1$ and $\Pi_2$ are {\em strongly equivalent\/} if 
$\Pi_1\cup\Pi$ and $\Pi_2\cup\Pi$ are equivalent for any program $\Pi$ (\cite{LPV01}).  In particular, 
two rules $r_1$ and $r_2$ are {\em strongly equivalent\/} if $\Pi\cup\{r_1\}$ and $\Pi\cup\{r_2\}$ are equivalent for any program $\Pi$. 

An answer set of a GEDP is not always minimal, i.e., a program $\Pi$ may have two answer sets $S$ and $T$ 
such that $S\subset T$. This is in contrast with the case of EDPs where every answer set is minimal. 

\begin{example}\rm 
Let $\Pi$ be the program:
\begin{eqnarray*}
&& p\,;\, not\,q\leftarrow,\\
&& q\,;\,not\,p\leftarrow. 
\end{eqnarray*}
Then $\Pi$ has two answer sets $\emptyset$ and $\{p,q\}$. 
\end{example}

By definition, a contradictory GEDP has exactly one answer set $Lit$, while 
a consistent GEDP may have the answer set $Lit$. 

\begin{example}\rm 
Let $\Pi$ be the program:
\begin{eqnarray*}
&& p\,;\, not\,p\leftarrow,\\
&& \neg\,p\leftarrow p. 
\end{eqnarray*}
Then $\Pi$ has two answer sets $\emptyset$ and $Lit$. 
\end{example}

In EDPs, on the other hand, no consistent program has the answer set $Lit$, and 
every contradictory program has exactly one answer set $Lit$ (\cite{GL91}). 

Suppose a rule $r$ such that $head^+(r)=\emptyset$: 
\begin{equation}\label{ic-rule}
not\,L_{k+1}\,;\cdots ;\,not\,L_l \leftarrow\; L_{l+1},\,\ldots,\,L_m,\,not\,L_{m+1},\,\ldots,\,not\,L_n. 
\end{equation}
Define a rule $\eta(r)$ of the form:
\begin{equation}\label{shift-rule}
\leftarrow\; L_{k+1},\,\ldots,\,L_l, L_{l+1},\,\ldots,\,L_m,\,not\,L_{m+1},\,\ldots,\,not\,L_n
\end{equation}
that is obtained by shifting ``$not\,L_{k+1}\,;\cdots ;\,not\,L_l $" in $head^-(r)$ to ``$L_{k+1},\,\ldots,\,L_l$" in $body^+(\eta(r))$. 
The two rules~(\ref{ic-rule}) and~(\ref{shift-rule}) are strongly equivalent under the answer set semantics. 

\begin{proposition}[\cite{IS98}]\label{ic-trans}\em
Let $\Pi$ be a program and $\Phi=\{\,r\mid r\in\Pi\;\;\mbox{and}\;\; head^+(r)=\emptyset\}$.  
Also let $\Pi'=(\Pi\setminus \Phi)\cup \{\, \eta(r)\mid r\in\Phi\,\}$. 
Then $\Pi$ and $\Pi'$ have the same answer sets. 
\end{proposition}

\begin{proposition}\em \label{deletion-prop}
Let $\Pi$ be a program and 
$\Psi=\{\,r\mid r\in\Pi,\; head^+(r)=\emptyset\;\mbox{and}\; head^-(r)\cap body^-(r)\neq\emptyset\}$. 
Then, $\Pi$ and $\Pi\setminus\Psi$ have the same answer sets. 
\begin{proof}\rm
By Proposition~\ref{ic-trans}, every rule~(\ref{ic-rule}) in $\Pi$ is transformed to a strongly equivalent constraint~(\ref{shift-rule}). 
When $head^-(r)\cap body^-(r)\neq\emptyset$ for some $r\in\Phi$, 
$\{ L_{k+1},\ldots,L_l\}\cap \{L_{m+1},\ldots,L_n\}\neq\emptyset$ in $\eta(r)$ of the form~(\ref{shift-rule}). 
Then, $\{ L_{k+1},\ldots,L_l\}\subseteq S$ implies $\{L_{m+1},\ldots,L_n\}\cap S\neq\emptyset$ for any set $S$, 
and the constraint $\eta(r)$ is satisfied by any answer set. 
Hence, $\Psi$ is removed from $\Pi$ and the result follows. 
\end{proof}
\end{proposition}

\begin{example}\rm
For any program $\Pi$, 
\[  \Pi \cup \{\, not\,p\,\leftarrow\, q, not\,p\,\} \] is equivalent to the following program (Proposition~\ref{ic-trans}): 
\[  \Pi \cup \{\, \leftarrow p, q, not\,p\,\}, \] 
which is further simplified to $\Pi$  (Proposition~\ref{deletion-prop}). 
\end{example}

\begin{proposition}\em \label{no-lit-prop}
Let $\Pi$ be a not-free GEDP. 
If there is a constraint in $\Pi$, then $\Pi$ is not contradictory. 
\begin{proof}\rm
If there is a constraint ``$\leftarrow L_1,\ldots,L_m$" in $\Pi$, it is included in $\Pi^{Lit}$. 
Since $Lit$ does not satisfy the constraint, it does not become the answer set of $\Pi^{Lit}$. 
Hence, $\Pi$ is not contradictory. 
\end{proof}
\end{proposition}


\section{Human Conditional Reasoning in ASP}  \label{sec:3}

ASP computes answer sets by deduction that is reasoning by AA. 
In this section, we present methods for reasoning by AC, DC, and DA in ASP. 

\subsection{AC Completion}

We first introduce a framework for reasoning by {\em affirming the consequent\/} (AC) in ASP. 
In GEDPs, a conditional sentence ``$\phi\Rightarrow \psi$" ({\em if\/} $\phi$ {\em then\/} $\psi$) is represented by the rule 
``$\psi~\leftarrow~\phi$"  
where $\psi$ is a disjunction ``$L_1\,;\,\cdots\,;\, L_k\, ;\, not\,L_{k+1}\,;\cdots ;\,not\,L_l$" and $\phi$ is a conjunction 
``$L_{l+1},\,\ldots,\,L_m,\,not\,L_{m+1},\,\ldots,\,not\,L_n$". 
To realize reasoning backward from $\psi$ to $\phi$, we extend a program $\Pi$ by introducing new rules. 

\begin{definition}[AC completion]\rm \label{df-accomp}
Let $\Pi$ be a program and $r\in\Pi$ a rule of the form:
\begin{eqnarray*}
&& L_1\,;\,\cdots\,;\, L_k\, ;\, not\,L_{k+1}\,;\cdots ;\,not\,L_l\\
&& \quad\qquad \leftarrow\; L_{l+1},\,\ldots,\,L_m,\,not\,L_{m+1},\,\ldots,\,not\,L_n.\;\;
\end{eqnarray*}
\begin{enumerate}\setlength{\itemsep}{+1mm}\setlength{\itemindent}{10pt}
\item For each disjunct in $head^+(r)$ and $head^-(r)$, converse the implication: 
\begin{eqnarray}
&& L_{l+1},\,\ldots,\,L_m,\, not\,L_{m+1},\,\ldots,\,not\,L_n \;\leftarrow\; L_j \;\;\; (1\leq j\leq k),  \label{rev-1}\\
&& L_{l+1},\,\ldots,\,L_m,\, not\,L_{m+1},\,\ldots,\,not\,L_n \;\leftarrow\; not\,L_j\;\;\; (k+1\leq j\leq l). \label{rev-2}
\end{eqnarray}
In (\ref{rev-1}) and (\ref{rev-2}), the conjunction ``$L_{l+1},\,\ldots,\,L_m,\, not\,L_{m+1},\,\ldots,\,not\,L_n$" appears on the 
left of ``$\leftarrow$".  
The produced (\ref{rev-1}) (resp.\ (\ref{rev-2})) is considered an abbreviation of the collection of $(n-l)$ rules:  
 $(L_{l+1}\leftarrow L_j),\ldots, (not\,L_n\leftarrow L_j)$ (resp.\ $(L_{l+1}\leftarrow not\,L_j),\ldots, (not\,L_n\leftarrow not\,L_j)$)\footnote{We often use the parenthesis `()' to improve the readability.}, 
hence we abuse the term `rule' and call (\ref{rev-1}) or (\ref{rev-2}) a rule. 
In particular, (\ref{rev-1}) is not produced if $head^+(r)=\emptyset$ or $body^+(r)=body^-(r)=\emptyset$; and 
(\ref{rev-2}) is not produced if $head^-(r)=\emptyset$ or $body^+(r)=body^-(r)=\emptyset$. 
The set of all rules~(\ref{rev-1}) and~(\ref{rev-2}) is denoted as $conv(r)$. 
\item Define 
\begin{eqnarray*} 
ac(\Pi)&\!=\!& \{\; \Sigma_1\,;\,\cdots\, ;\,\Sigma_p \;\leftarrow \; \ell_j  \;\mid\\
&& \quad \Sigma_i\leftarrow \ell_j \;\, (1\le i\le p)\;\;\mbox{is in}\;\; \bigcup_{r\in\Pi}\, conv(r)\, \} 
\end{eqnarray*}
where each $\Sigma_i$ $(1\le i\le p)$ is a conjunction of literals and NAF-literals,  
and $\ell_j$ is either a literal $L_j$ $(1\le j\le k)$ or an NAF-literal $not\,L_j$ $(k+1\le j\le l)$. 
\item The {\em AC completion\/} of $\Pi$ is defined as:
\[ AC(\Pi) = \Pi \;\cup\; ac(\Pi). \] 
\end{enumerate}
\end{definition}

(\ref{rev-1}) and~(\ref{rev-2}) in $conv(r)$ represent converse implications from the disjunction in the head of $r$ 
to the conjunction in the body of $r$.  
$ac(\Pi)$ collects rules ``$\Sigma_i\leftarrow \ell_j$" $(1\le i\le p)$ having the same (NAF-)literal $\ell_j$ 
on the right of ``$\leftarrow$", and constructs 
``$\Sigma_1\,;\,\cdots\, ;\,\Sigma_p \;\leftarrow \; \ell_j$", which we call an {\em extended rule}. 
Introducing $ac(\Pi)$ to $\Pi$ realizes reasoning by AC in $\Pi$. 

The set $ac(\Pi)$ contains an extended rule having a disjunction of conjunctions in its head, while it is transformed to rules of a GEDP. 
That is, the extended rule:
\[ (\ell^1_1,\ldots,\ell^1_{m_1})\, ;\, \cdots\, ; \, (\ell^p_1,\ldots,\ell^p_{m_p})\leftarrow \ell_j \]
is identified with the set of $(m_1\times\cdots\times m_p)$ rules of the form:
\[ \ell^1_{i_1}\, ;\, \cdots\, ; \, \ell^p_{i_p}\leftarrow \ell_j \;\;\;\;\; (1\le i_k\le m_k;\, 1\le k\le  p).\]
By this fact, $AC(\Pi)$ is viewed as a GEDP and we do not distinguish extended rules and rules of a GEDP hereafter. 
The semantics of $AC(\Pi)$ is defined by its answer sets. 

\begin{example}\rm 
Let $\Pi$ be the program: 
\begin{eqnarray*}
&& p\,;\, not\,q\leftarrow r,\, not\,s, \\
&& p\leftarrow q.
\end{eqnarray*}
Then $ac(\Pi)$ becomes 
\begin{eqnarray*}
&& (r,\, not\,s)\,;\, q\leftarrow p,\\
&& r,\, not\,s\leftarrow not\,q
\end{eqnarray*}
where the first rule ``$(r,\, not\,s)\,;\, q\leftarrow p$" is identified with 
\begin{eqnarray*}
&& r\,;\, q\leftarrow p,\\
&& not\,s\,;\, q\leftarrow p,
\end{eqnarray*}
and the second rule ``$r,\, not\,s\leftarrow not\,q$" is identified with 
\begin{eqnarray*}
&& r\leftarrow not\,q,\\
&& not\,s\leftarrow not\,q. 
\end{eqnarray*}
Then, $AC(\Pi)\cup \{p\leftarrow\}$ has two answer sets $\{p,q\}$ and $\{p,r\}$. 
\end{example}

By definition, if there is more than one rule in $\Pi$ having the same (NAF-)literal in the heads, they are collected to produce a 
single converse rule in $ac(\Pi)$.  For instance, $\Pi=\{\, p\leftarrow~q,\;\; p\leftarrow~r\,\}$ produces 
$ac(\Pi)=\{\, q\,;\,r\leftarrow p\,\}$ but not $\Lambda=\{\, q\leftarrow p,\;\; r\leftarrow p\,\}$. 
Then, $AC(\Pi)\cup\{ p\leftarrow\}$ has two answer sets $\{p,q\}$ and $\{p,r\}$.  
Suppose that the new fact ``$\neg\, q\leftarrow$" is added to $\Pi$. 
Put $\Pi'=\Pi\cup\{\,\neg\, q\leftarrow\,\}$.  Then $AC(\Pi')\cup\{ p\leftarrow\}$ has the answer set $\{p,r\}$, which represents 
the result of AC reasoning in $\Pi'$.  If $\Lambda$ is used instead of $ac(\Pi)$, however, $\Pi'\cup\Lambda\cup\{ p\leftarrow\}$ 
has the answer set $Lit$. The result is too strong because $r$ is consistently inferred from 
$\Pi'\cup\{ p\leftarrow\}$ by AC reasoning. 
As a concrete example, put $p=wet\mbox{-}grass$, $q=rain$, and $r=sprinkler\mbox{-}on$. 
Then $AC(\Pi')\cup\{\, wet\mbox{-}grass\leftarrow\,\}$ has the answer set $\{\,wet\mbox{-}grass, \neg\, rain, sprinkler\mbox{-}on\,\}$, while $\Pi'\cup\Lambda\cup\{\, wet\mbox{-}grass\leftarrow\,\}$ has the answer set $Lit$. 
AC completion derives an antecedent from a consequent, but it does not derive negation of antecedent by its nature. 
For instance, given $\Pi=\{\, p\,;\,q\leftarrow r,\;\;\; p\leftarrow\,\}$, $AC(\Pi)\models_s r$ but $AC(\Pi)\not\models_c \neg\,q$.  

Note that in Definition~\ref{df-accomp} the converse of constraints and facts are not produced. 
When $head^+(r)=head^-(r)=\emptyset$, $r$ is considered a rule with $\mathit{false}$ in the head, then 
$(\ref{rev-1})$ and $(\ref{rev-2})$ become 
\[ L_{l+1},\,\ldots,\,L_m,\, not\,L_{m+1},\,\ldots,\,not\,L_n \;\leftarrow \mathit{false}\] 
which has no effect as a rule. 
On the other hand, when $body^+(r)=body^-(r)=\emptyset$, $r$ is considered a rule with $\mathit{true}$ in the body, then 
$(\ref{rev-1})$ and $(\ref{rev-2})$ respectively become 
\[ true\leftarrow L_j\; (1\leq j\leq k)\quad\mbox{and}\quad true\leftarrow not\,L_j\; (k+1\leq j\leq l).\] 
We do not include this type of rules for constructing $ac(\Pi)$ because it would disable AC reasoning. 
For instance, transform $\Pi=\{\, p\leftarrow q,\;\;\; p\leftarrow\,\}$ to $\Pi'=\Pi\cup\{ q\,; true\leftarrow p\}$. 
Then $\{p\}$ is the minimal set satisfying $\Pi'$, and $q$ is not included in the answer set of $\Pi'$. 
With this reason, constraints and facts are not completed at the first step of Definition~\ref{df-accomp}. 

The result of AC completion is syntax-dependent in general. 
That is, two (strongly) equivalent programs may produce different AC completions.   

\begin{example}\rm 
Let $\Pi_1=\{\, not\,p\leftarrow q\,\}$ and $\Pi_2=\{\, \leftarrow p,q \,\}$. 
By Proposition~\ref{ic-trans}, $\Pi_1$ and $\Pi_2$ are equivalent, but 
$AC(\Pi_1)=\Pi_1\cup \{\, q\leftarrow not\,p\,\}$ and $AC(\Pi_2)=\Pi_2$. 
As a result, $AC(\Pi_1)$ has the answer set  $\{q\}$ while $AC(\Pi_2)$ has the answer set $\emptyset$. 
\end{example}

In the above example, ``$not\,p\leftarrow q$" is a conditional sentence which is subject to AC inference, while 
``$\leftarrow p,q$" is a constraint which is not subject to AC inference by definition. 
For instance, given the conditional sentence ``if it is sunny, the grass is not wet" and the fact ``the grass is not wet", 
people would infer ``it is sunny" by AC inference. 
On the other hand, given the constraint ``it does not happen that wet-grass and sunny-weather at the same time" and 
the fact ``the grass is not wet", the number of people who infer ``it is sunny" by AC would be smaller because the cause-effect relation 
between ``sunny" and ``not wet" is not explicitly expressed in the constraint. 

Reasoning by AC is nonmonotonic in the sense that $\Pi\models_c L$ (or $\Pi\models_s L$)
does not imply $AC(\Pi)\models_c L$ (or $AC(\Pi)\models_s L$) in general. 

\begin{example}\rm \label{nonmon-ex}
The program $\Pi=\{\, p\leftarrow not\,q,\;\; r\leftarrow q,\;\; r\leftarrow \,\}$ has the answer set $\{p,r\}$, while 
$AC(\Pi)=\Pi\cup\{\, not\,q\leftarrow p,\;\;  q\leftarrow r\,\}$ has the answer set $\{q,r\}$. 
\end{example}
 
In Example~\ref{nonmon-ex}, reasoning by AC produces $q$ which blocks deriving $p$ using the first rule in $\Pi$.  
As a concrete example, an online-meeting is held on time if no network trouble arises. 
However, it turns that the web browser is unconnected and one suspects that there is some trouble on the network. 
Put $p$=``online-meeting is held on time", $q$=``network trouble",  $r$=``the web browser is unconnected". 
In this case, one may withdraw the conclusion $p$ after knowing $r$.  
As such, additional rules $ac(\Pi)$ may change the results of $\Pi$. 
One can see the effect of AC reasoning in a program $\Pi$ by comparing answer sets of $\Pi$ and $AC(\Pi)$. 

A consistent program $\Pi$ may produce an inconsistent $AC(\Pi)$. 
In converse, an inconsistent $\Pi$ may produce a consistent $AC(\Pi)$. 

\begin{example}\rm \label{inconsistent-ac-ex}
$\Pi_1=\{\, p\leftarrow \neg\, p,\;\;\; p\leftarrow\,\}$ is consistent, but 
$AC(\Pi_1)=\Pi_1\cup \{\, \neg\,p\leftarrow p\,\}$ is contradictory. 
$\Pi_2=\{\, \leftarrow not\,p,\;\;\; q\leftarrow p,\;\;\; q\leftarrow\,\}$ is incoherent, but 
$AC(\Pi_2)=\Pi_2\cup \{\, p\leftarrow q\,\}$ is consistent. 
\end{example}

A sufficient condition for the consistency of $AC(\Pi)$ is given below. 

\begin{proposition}\em \label{ac-pdp-prop}
If a PDP $\Pi$ contains no constraint, then $AC(\Pi)$ is consistent. 
Moreover, for any answer set $S$ of $\Pi$, there is an answer set $T$ of $AC(\Pi)$ such that $S\subseteq T$. 
\begin{proof}\rm
A PDP $\Pi$ contains no NAF-literal and every literal in $\Pi$ is a positive literal (or an atom). 
Then $ac(\Pi)$ is the set of rules $(\Sigma_1;\cdots;\Sigma_p\leftarrow A_j)$ 
where $A_j$ is an atom and $\Sigma_i$ is a conjunction of atoms. 
When $\Pi$ contains no constraint, the additional rules in $ac(\Pi)$ do not cause inconsistency in 
$AC(\Pi)=\Pi\cup ac(\Pi)$.  
Suppose that $S$ is an answer set (or a minimal model) of $\Pi$.  Then $S$ is a minimal set satisfying all rules in $\Pi$. 
Let $U=\{\, A\, \mid\, A\in head^+(r)\;\mbox{for any rule}\; r\in ac(\Pi)\;\mbox{such that}\; body^+(r)\subseteq S\,\}$. 
Then there is an answer set $T$ of $AC(\Pi)$ such that $T=S\cup V$ where $V\subseteq U$. 
Hence the result holds. 
\end{proof}
\end{proposition}

\begin{proposition}\em \label{ac-contra-prop}
If a program $\Pi$ has the answer set $Lit$, then $AC(\Pi)$ has the answer set $Lit$. 
\begin{proof}\rm 
If $\Pi$ has the answer set $Lit$, the reduct $\Pi^{Lit}$ has the answer set $Lit$. 
By definition, $AC(\Pi)^{Lit}=\Pi^{Lit} \cup ac(\Pi)^{Lit}$ where $ac(\Pi)^{Lit}$ is the reduct of $ac(\Pi)$ by $Lit$. 
Introducing $not$-free rules in $ac(\Pi)^{Lit}$ does not change the answer set $Lit$ of $\Pi^{Lit}$. 
Then, $Lit$ is the answer set of $AC(\Pi)^{Lit}$ and the result follows. 
\end{proof}
\end{proposition}

\subsection{DC Completion}

We next introduce a framework for reasoning by denying the consequent (DC) in ASP.  
There are two ways for negating a literal -- one is using explicit negation and the other is using default negation. 
Accordingly, there are two ways of completing a program for the purpose of reasoning by DC. 

\begin{definition}[DC completion]\rm \label{def-dc-comp}
Let $\Pi$ be a program.  For each rule $r\in\Pi$ of the form: 
\begin{eqnarray*}
&& L_1\,;\,\cdots\,;\, L_k\, ;\, not\,L_{k+1}\,;\cdots ;\,not\,L_l\\
&& \qquad \leftarrow\; L_{l+1},\,\ldots,\,L_m,\,not\,L_{m+1},\,\ldots,\,not\,L_n\;\;
\end{eqnarray*}
define $wdc(r)$ as the rule: 
\begin{equation}\label{wdc}
not\,L_{l+1};\cdots ; not\,L_m \,;\, L_{m+1}; \cdots; L_n 
\leftarrow  not\, L_1,\ldots, not\,L_k,\, L_{k+1},\ldots, L_l 
\end{equation}
and define $sdc(r)$ as the rule: 
\begin{equation}\label{sdc}
\neg\,L_{l+1};\cdots ; \neg\,L_m \,;\, L_{m+1}; \cdots; L_n
\leftarrow  \neg\, L_1,\ldots, \neg\,L_k,\, L_{k+1},\ldots, L_l.  
\end{equation}
In particular, (\ref{wdc}) or (\ref{sdc}) becomes a fact if $head^+(r)=head^-(r)=\emptyset$; and 
it becomes a constraint if $body^+(r)=body^-(r)=\emptyset$. 
The {\em weak DC completion\/} and the {\em strong DC completion\/} of $\Pi$ are respectively defined as:
\begin{eqnarray*}
WDC(\Pi) &=& \Pi \;\cup\; \{\, wdc(r) \,\mid\, r\in \Pi \,\}, \\
SDC(\Pi) &=& \Pi \;\cup\; \{\, sdc(r) \,\mid\, r\in \Pi \,\}. 
\end{eqnarray*}
\end{definition}

By definition, $WDC(\Pi)$ and $SDC(\Pi)$ introduce contrapositive rules in two different ways. 
In (\ref{wdc}), literals $L_i$ $(1\leq i\leq k;\,l+1\leq i\leq m)$ are negated using default negation $not$ 
and NAF-literals $not\,L_j$ $(k+1\leq i\leq l;\,m+1\leq i\leq n)$ are converted to $L_j$. 
In (\ref{sdc}), on the other hand, 
literals $L_i$ $(1\leq i\leq k;\,l+1\leq i\leq m)$ are negated using explicit negation  $\neg$ 
and NAF-literals $not\,L_j$ $(k+1\leq i\leq l;\,m+1\leq i\leq n)$ are converted to $L_j$. 
$WDC(\Pi)$ and $SDC(\Pi)$ are GEDPs and their semantics are defined by their answer sets. 
In particular, $SDC(\Pi)$ becomes an EDP if $\Pi$ is an EDP. 

Note that contraposition of facts or constraints is produced in WDC/SDC. 
For instance, the fact ``$p\leftarrow$" produces the constraint ``$\leftarrow not\,p$" by WDC and ``$\leftarrow \neg p$" by SDC. 
The fact ``$not\,p\leftarrow$" produces the constraint ``$\leftarrow p$" by WDC and SDC. 
On the other side, the constraint ``$\leftarrow p$" produces the fact ``$not\,p\leftarrow$" by WDC and ``$\neg p\leftarrow$" by SDC. 
The constraint ``$\leftarrow not\,p$" produces the fact ``$p\leftarrow$" by WDC and SDC. 

The WDC and SDC produce different results in general. 

\begin{example}\label{wdc-sdc-ex}\rm 
Given $\Pi=\{\,p\leftarrow not\, q\,\}$, it becomes 
\begin{eqnarray*}
 WDC(\Pi)&=&\{\, p\leftarrow not\,q, \quad  q\leftarrow not\,p\,\},\\
 SDC(\Pi)&=&\{\,  p \leftarrow not\,q, \quad q\leftarrow \neg\,p\,\}.
\end{eqnarray*}
Then $WDC(\Pi)$ has two answer sets $\{p\}$ and $\{q\}$, while $SDC(\Pi)$ has the single answer set $\{ p\}$. 
\end{example}

Example~\ref{wdc-sdc-ex} shows that WDC is nonmonotonic as $\Pi\models_s p$ but $WDC(\Pi)\not\models_s p$. 
SDC is also nonmonotonic (see Example~\ref{ex-sdc-inc}). 
The result of DC completion is syntax-dependent in general. 

\begin{example}\rm 
Let $\Pi_1=\{\, not\,p\leftarrow q\,\}$ and $\Pi_2=\{\, \leftarrow p,q \,\}$ where $\Pi_1$ and $\Pi_2$ are equivalent 
(Proposition~\ref{ic-trans}).  Then, 
$SDC(\Pi_1)=\Pi_1\cup \{\, \neg\,q\leftarrow p\,\}$ and $SDC(\Pi_2)=\Pi_2\cup\{\,\neg\,p\,; \neg\,q\leftarrow\,\}$. 
As a result, $SDC(\Pi_1)$ has the answer set  $\emptyset$, while $SDC(\Pi_2)$ has two answer sets $\{\neg\,p\}$ and $\{\neg\,q\}$. 
\end{example}

WDC preserves the consistency of the original program. 

\begin{proposition}\em \label{wdc-consist}
If a program $\Pi$ has a consistent answer set $S$, then $S$ is an answer set of $WDC(\Pi)$. 
\begin{proof}\rm
Suppose a consistent answer set $S$ of $\Pi$. Then, for any rule 
($L_1\, ;\,\cdots\, ;\, L_k \leftarrow L_{l+1},\,\ldots,\,L_m$) in $\Pi^S$, 
either $\{L_1,\ldots,L_k\}\cap S\neq\emptyset$ or $\{L_{l+1},\ldots,L_m\}\not\subseteq S$. 
In each case, the rule~(\ref{wdc}) is eliminated in $\Pi^S$.  Then, $S$ is an answer set of $WDC(\Pi)^S$, hence 
the result holds. 
\end{proof}
\end{proposition}

The converse of Proposition~\ref{wdc-consist} does not hold in general. 

\begin{example}\rm 
The program $\Pi=\{\, \leftarrow not\,p \,\}$ has no answer set, while 
$WDC(\Pi)=\{\, \leftarrow~not\,p,\;\; p\leftarrow \,\}$ has the answer set $\{p\}$. 
\end{example}

For $SDC(\Pi)$, the next result holds. 

\begin{proposition}\em \label{sdc-consist}
Let $\Pi$ be a consistent program such that every constraint in $\Pi$ is $not$-free   
(i.e., $head^+(r)=head^-(r)=\emptyset$ implies $body^-(r)=\emptyset$ for any $r\in\Pi$). 
Then $SDC(\Pi)$ does not have the answer set $Lit$. 
\begin{proof}\rm
Consider $SDC(\Pi)^{Lit}=(\Pi \;\cup\; \{\, sdc(r) \,\mid\, r\in \Pi \,\})^{Lit}=\Pi^{Lit}\;\cup\; \{\, sdc(r) \,\mid\, r\in \Pi \,\}$. \\
(a) If there is a constraint $r\in \Pi$ such that $head^+(r)=head^-(r)=\emptyset$, then $body^-(r)=\emptyset$ by the assumption. 
Then $r\in\Pi^{Lit}$ and $SDC(\Pi)^{Lit}$ does not have the answer set $Lit$ (Proposition~\ref{no-lit-prop}). 
(b) Else if there is no constraint in $\Pi$, then $\{\, sdc(r) \,\mid\, r\in \Pi \,\}$ contains no fact by Def.~\ref{def-dc-comp}. 
Consider two cases. 
(i) When there is a fact $r\in\Pi$, $sdc(r)$ becomes a not-free constraint by Def.~\ref{def-dc-comp}. 
As $SDC(\Pi)^{Lit}$ contains this constraint, it does not have the answer set $Lit$ (Proposition~\ref{no-lit-prop}). 
(ii) When there is no fact in $\Pi$, both $\Pi^{Lit}$ and $\{\, sdc(r) \,\mid\, r\in \Pi \,\}$ contain no fact, so 
$SDC(\Pi)^{Lit}$ contains no fact.  In this case, no literal is deduced in $SDC(\Pi)^{Lit}$ and it does not have the answer set $Lit$. 
By (a) and (b), $SDC(\Pi)^{Lit}$ does not have the answer set $Lit$.  Hence, the result follows. 
\end{proof}
\end{proposition}

A program $\Pi$ satisfying the condition of Proposition~\ref{sdc-consist} may produce an incoherent $SDC(\Pi)$. 

\begin{example}\label{ex-sdc-inc}\rm 
The program 
$\Pi = \{\, p\leftarrow q,\;\;\;  p\leftarrow \neg q,\;\;\; \neg\,p\leftarrow\, \}$ has the answer set $\{\neg\,p\}$, 
but 
\[ SDC(\Pi)=\Pi \cup \{\,\neg\,q\leftarrow \neg\,p,\;\;\; q\leftarrow \neg\,p,\;\;\; \leftarrow p\,\} \]
is incoherent.
\end{example}

\begin{proposition}\em
If a program $\Pi$ has the answer set $Lit$, then both $WDC(\Pi)$ and $SDC(\Pi)$ have the answer set $Lit$. 
In particular, if\, $\Pi$ is a contradictory EDP, then $SDC(\Pi)$ is contradictory. 
\begin{proof}\rm 
The proof is similar to Proposition~\ref{ac-contra-prop}. 
In particular, if $\Pi$ is an EDP, then $SDC(\Pi)$ is an EDP and the result holds. 
\end{proof}
\end{proposition}

Note that in GEDPs contraposition of a rule does not hold in general. 
Thus, the program $\Pi=\{\,  p\leftarrow q,\;\; \neg\,p\leftarrow \,\}$ does not deduce $\neg\,q$. 
$SDC$ completes the program as 
$SDC(\Pi)=\Pi\cup \{\,  \neg\,q\leftarrow \neg\,p,\;\; \leftarrow p\,\}$ and makes $\neg\,q$ deducible. 
In this sense, SDC has the effect of making explicit negation closer to classical negation in GEDP. 

\subsection{DA Completion}

As a third extension, we introduce a framework for reasoning by denying the antecedent (DA) in ASP.  
As in the case of DC completion, two different ways of completion are considered depending on the choice of negation. 

\begin{definition}[weak DA completion]\rm \label{wda-def}
Let $\Pi$ be a program and $r\in\Pi$ a rule of the form: 
\begin{eqnarray*}
&& L_1\,;\,\cdots\,;\, L_k\, ;\, not\,L_{k+1}\,;\cdots ;\,not\,L_l\\
&& \quad\qquad\leftarrow\; L_{l+1},\,\ldots,\,L_m,\,not\,L_{m+1},\,\ldots,\,not\,L_n.
\end{eqnarray*}
\begin{enumerate}\setlength{\itemsep}{+1mm}\setlength{\itemindent}{10pt}
\item For each disjunct in $head^+(r)$ and $head^-(r)$, inverse the implication: 
\begin{eqnarray}
not\,L_i &\leftarrow& not\,L_{l+1}\,;\,\cdots\, ;\, not\,L_m\,;\, L_{m+1}\,;\cdots\, ;\, L_n\;\; (1\leq i\leq k),  \label{wda-1}\\
L_i &\leftarrow& not\,L_{l+1}\,;\,\cdots\, ;\, not\,L_m\,;\, L_{m+1}\,;\,\cdots\, ;\, L_n\;\; (k+1\leq i\leq l). \;\;\;\;\;\label{wda-2}
\end{eqnarray}
In (\ref{wda-1}) and (\ref{wda-2}), the disjunction ``$not\,L_{l+1}\,;\,\cdots\, ;\, not\,L_m\,;\, L_{m+1}\,;\,\cdots\, ;\, L_n$"  
appears on the right of ``$\leftarrow$". 
The produced (\ref{wda-1}) (resp.\ (\ref{wda-2})) is considered an abbreviation of the collection of $(n-l)$ rules: 
$(not\,L_i \leftarrow not\,L_{l+1}),\ldots, (not\,L_i \leftarrow L_n)$ (resp.\ 
$(L_i \leftarrow not\,L_{l+1}),\ldots, (L_i \leftarrow L_n)$), 
hence we abuse the term `rule' and call (\ref{wda-1}) or (\ref{wda-2}) a rule. 
In particular, (\ref{wda-1}) is not produced if $head^+(r)=\emptyset$ or $body^+(r)=body^-(r)=\emptyset$; and 
(\ref{wda-2}) is not produced  if $head^-(r)=\emptyset$ or $body^+(r)=body^-(r)=\emptyset$. 
The set of rules~(\ref{wda-1})--(\ref{wda-2}) is denoted as $winv(r)$. 

\item Define 
\begin{eqnarray*}
wda(\Pi) &\!=\!& \{\, \ell_i \leftarrow \Gamma_1,\ldots, \Gamma_p \,\mid \\
&& \quad \ell_i\leftarrow \Gamma_j\; (1\leq j\leq p)\;\;\mbox{is in}\;\; \bigcup_{r\in\Pi}\; winv(r)\,\}
\end{eqnarray*}
where $\ell_i$ is either a literal  $L_i$ $(k+1\le i\le l)$ or an NAF-literal  $not\,L_i$ $(1\le i\le k)$, and 
each $\Gamma_j$ $(1\le j\le p)$ is a disjunction of literals and NAF-literals.   

\item The {\em weak DA completion\/} of $\Pi$ is defined as:
\[ WDA(\Pi)=\Pi \;\cup\; wda(\Pi).\]
\end{enumerate}
\end{definition}

(\ref{wda-1}) and  (\ref{wda-2}) in $winv(r)$ represent inverse implication from the (default) negation of the conjunction in the body of $r$ to 
the (default) negation of the disjunction in the head of $r$.  
$wda(\Pi)$ collects rules ``$\ell_i\leftarrow \Gamma_j$" $(1\leq j\leq p)$ having the same 
(NAF-)literal $\ell_i$ on the left of ``$\leftarrow$", and constructs ``$\ell_i \leftarrow \Gamma_1,\ldots, \Gamma_p$", which we call an 
{\em extended rule}.  Introducing $wda(\Pi)$ to $\Pi$ realizes reasoning by weak DA. 
An extended rule has a conjunction of disjunctions in its body, while it is transformed to rules of a GEDP as the case of AC completion. 
That is, the extended rule: 
\[ \ell_i \leftarrow (\ell^1_1\,;\,\cdots\,;\,\ell^1_{m_1})\, , \ldots,\, (\ell^p_1\,;\,\cdots\,;\,\ell^p_{m_p})\]
is identified with the set of $(m_1\times\cdots\times m_p)$ rules of the form:
\[ \ell_i \leftarrow \ell^1_{j_1} ,\, \ldots,\,  \ell^p_{j_p} \;\;\;\;\; (1\le j_k\le m_k;\, 1\le k\le  p).\]
By this fact, $WDA(\Pi)$ is viewed as a GEDP and we do not distinguish extended rules and rules of a GEDP hereafter. 
The semantics of $WDA(\Pi)$ is defined by its answer sets. 

\begin{example}\rm 
Let $\Pi$ be the program: 
\begin{eqnarray*}
&& p\,;\,q\leftarrow r,\, not\,s,\\
&& q\,;\,not\,r\leftarrow t,\\
&& s\leftarrow.
\end{eqnarray*}
Then $wda(\Pi)$ becomes 
\begin{eqnarray*}
&& not\,p\leftarrow not\,r\, ;\, s,\\
&& not\,q\leftarrow (not\,r\,;\, s),\, not\,t,\\
&& r\leftarrow not\,t
\end{eqnarray*}
where the first rule ``$not\,p\leftarrow not\,r\, ;\, s$" is identified with 
\begin{eqnarray*}
&& not\,p\leftarrow not\,r,\\
&& not\,p\leftarrow s,
\end{eqnarray*}
and the second rule ``$not\,q\leftarrow (not\,r\,;\, s),\, not\,t$" is identified with 
\begin{eqnarray*}
&& not\,q\leftarrow not\,r,\,not\,t,\\
&& not\,q\leftarrow s,\, not\,t.
\end{eqnarray*}
Then, $WDA(\Pi)$ has the answer set $\{ s, r\}$. 
\end{example}

As in the case of AC completion, if there is more than one rule having the same (NAF-)literal in the heads, 
they are collected to produce a single inverse rule.  For instance, $\Pi=\{\,p\leftarrow q,\;\;  p\leftarrow~r\,\}$  
produces $wda(\Pi)=\{\, not\,p\leftarrow not\,q,\,not\,r\,\}$ but not 
$\Lambda=\{\, not\,p\leftarrow not\,q,\;\; not\,p\leftarrow not\,r\,\}$. 
Suppose that the new  fact ``$r\leftarrow$" is added to $\Pi$. 
Put $\Pi'=\Pi\cup\{ r\leftarrow\}$. Then $WDA(\Pi')$ has the answer set $\{p,r\}$. 
If $\Lambda$ is used instead of $wda(\Pi)$, however, $\Pi'\cup\Lambda$ is  incoherent because the first rule of $\Lambda$ is 
not satisfied.  The result is too strong because $p$ is deduced by ``$p\leftarrow r$" and ``$r\leftarrow$", and it has no direct connection 
to DA inference in the first rule of $\Lambda$. Hence, we conclude $not\,p$ if both $q$ and $r$ are negated in $wda(\Pi)$. 

The strong DA completion is defined in a similar manner. 

\begin{definition}[strong DA completion]\rm \label{sda-def}
Let $\Pi$ be a program and $r\in\Pi$ a rule of the form: 
\begin{eqnarray*}
&& L_1\,;\,\cdots\,;\, L_k\, ;\, not\,L_{k+1}\,;\cdots ;\,not\,L_l\\
&& \qquad\leftarrow\; L_{l+1},\,\ldots,\,L_m,\,not\,L_{m+1},\,\ldots,\,not\,L_n. 
\end{eqnarray*}
\begin{enumerate}\setlength{\itemsep}{+1mm}\setlength{\itemindent}{10pt}
\item For each disjunct in $head^+(r)$ and $head^-(r)$, inverse the implication: 
\begin{eqnarray}
\neg\,L_i &\leftarrow& \neg\,L_{l+1}\,;\,\cdots\, ;\, \neg\,L_m\,;\, L_{m+1}\,;\cdots\, ;\, L_n\;\; (1\leq i\leq k),  \label{sda-1}\\
L_i &\leftarrow& \neg\,L_{l+1}\,;\,\cdots\, ;\, \neg\,L_m\,;\, L_{m+1}\,;\,\cdots\, ;\, L_n\;\; (k+1\leq i\leq l). \;\;\;\;\;\label{sda-2}
\end{eqnarray} 
As in the case of WDA, 
the produced (\ref{sda-1}) (resp.\ (\ref{sda-2})) is considered an abbreviation of the collection of $(n-l)$ rules: 
$(\neg\,L_i \leftarrow \neg\,L_{l+1}),\ldots, (\neg\,L_i \leftarrow L_n)$ (resp.\ 
$(L_i \leftarrow \neg\,L_{l+1}),\ldots, (L_i \leftarrow L_n)$), 
hence we call (\ref{sda-1}) or (\ref{sda-2}) a rule. 
In particular, (\ref{sda-1}) is not produced if $head^+(r)=\emptyset$ or $body^+(r)=body^-(r)=\emptyset$; and 
(\ref{sda-2}) is not produced if $head^-(r)=\emptyset$ or $body^+(r)=body^-(r)=\emptyset$. 
The set of rules~(\ref{sda-1})--(\ref{sda-2}) is denoted as $sinv(r)$. 

\item Define 
\begin{eqnarray*}
sda(\Pi) &\!=\!& \{\, \ell_i \leftarrow \Gamma_1,\ldots, \Gamma_p \,\mid \\
&& \quad \ell_i\leftarrow \Gamma_j\; (1\leq j\leq p)\;\;\mbox{is in}\;\; \bigcup_{r\in\Pi}\; sinv(r)\,\}
\end{eqnarray*}
where $\ell_i$ is either a literal  $L_i$ $(k+1\le i\le l)$ or $\neg\,L_i$ $(1\le i\le k)$, and 
each $\Gamma_j$ $(1\le j\le p)$ is a disjunction of positive/negative literals.   

\item The {\em strong DA completion\/} of $\Pi$ is defined as:
\[ SDA(\Pi)=\Pi \;\cup\; sda(\Pi).\]
\end{enumerate}
\end{definition}

As in the case of WDA, extended rules in $sda(\Pi)$ are transformed to rules of a GEDP. 
Then $SDA(\Pi)$ is viewed as a GEDP and its semantics is defined by its answer sets. 
In particular, $SDA(\Pi)$ becomes an EDP if $\Pi$ is an EDP. 

The result of DA completion is syntax-dependent in general. 

\begin{example}\rm 
Let $\Pi_1=\{\, not\,p\leftarrow q\,\}$ and $\Pi_2=\{\, \leftarrow p,q \,\}$ where $\Pi_1$ and $\Pi_2$ are equivalent 
(Proposition~\ref{ic-trans}).  Then, 
$WDA(\Pi_1)=\Pi_1\cup \{\, p\leftarrow not\,q\,\}$ and $WDA(\Pi_2)=\Pi_2$. 
As a result, $WDA(\Pi_1)$ has the answer set  $\{p\}$ while $WDA(\Pi_2)$ has the answer set $\emptyset$. 
\end{example}

Both WDA and SDA are nonmonotonic in general. 

\begin{example}\rm 
(1) $\Pi_1=\{\, p\leftarrow not\,q,\;\; not\,q\leftarrow p\,\}$ produces 
$WDA(\Pi_1)=\Pi_1\cup \{\, not\,p\leftarrow q,\;\; q\leftarrow not\,p\,\}$.  
Then, $\Pi_1\models_s p$ but $WDA(\Pi_1)\not\models_s p$. 
(2) $\Pi_2=\{\, p\leftarrow not\,\neg r,\;\; r\leftarrow not\,q,\;\; q\leftarrow\,\}$ produces 
$SDA(\Pi_2)=\Pi_2\cup \{\, \neg\,p\leftarrow \neg\,r,\;\; \neg\,r\leftarrow q\,\}$.  
Then, $\Pi_2\models_c p$ but $SDA(\Pi_2)\not\models_c p$. 
\end{example}

When a program is a consistent EDP, the WDA does not introduce a new answer set. 

\begin{proposition} \em
Let $\Pi$ be an EDP. If $S$ is a consistent answer set of\, $WDA(\Pi)$, then $S$ is an answer set of $\Pi$. 
\begin{proof}\rm
When $\Pi$ is an EDP, the rules~(\ref{wda-2}) are not included in $winv(r)$. 
By the rules~(\ref{wda-1}) in $winv(r)$, rules of the form ($not\,L_i\leftarrow \Gamma_1,\ldots, \Gamma_p$) 
are produced in $wda(\Pi)$, which are identified with the collection of rules 
($not\,L_i\leftarrow \ell_{j_1}^1,\ldots \ell_{j_p}^p$) where $\ell_{j_k}^p$ $(1\le k\le p)$ is an (NAF-)literal. 
These rules are converted into the strongly equivalent constraint 
($\leftarrow L_i,\ell_{j_1}^1,\ldots, \ell_{j_p}^p$) (Proposition~\ref{ic-trans}). 
 The additional constraints may eliminate answer sets of $\Pi$, but do not introduce new answer sets. 
 Thus, a consistent answer set $S$ of $WDA(\Pi)$ is also a consistent answer set of $\Pi$. 
\end{proof}
\end{proposition}

\begin{proposition}\em
If a program $\Pi$ has the answer set $Lit$, then both $WDA(\Pi)$ and $SDA(\Pi)$ have the answer set $Lit$. 
In particular, if\, $\Pi$ is a contradictory EDP, then $SDA(\Pi)$ is contradictory. 
\begin{proof}\rm 
The proof is similar to Proposition~\ref{ac-contra-prop}. 
In particular, if $\Pi$ is an EDP, then $SDA(\Pi)$ is an EDP and the result holds. 
\end{proof}
\end{proposition}

As in the case of AC, a consistent program $\Pi$ may produce an inconsistent $WDA(\Pi)$ or $SDA(\Pi)$. 
In converse, an incoherent $\Pi$ may produce a consistent $WDA(\Pi)$ or $SDA(\Pi)$. 

\begin{example}\label{inconsistent-da-ex}\rm
\begin{description}
\item[](1) $\Pi_1=\{\, not\, p\leftarrow p\,\}$, which is equivalent to $\{\,\leftarrow p\,\}$ (Proposition~\ref{ic-trans}), is consistent, 
but $WDA(\Pi_1)=\Pi_1\cup\{\, p\leftarrow not\,p\,\}$ is incoherent. 
\item[](2) $\Pi_2=\{\,\neg\,p\leftarrow p,\;\; \neg\,p\leftarrow\,\}$ is consistent, but $SDA(\Pi_2)\!=\Pi_2\cup\{\, p\leftarrow \neg\,p\,\}$ is contradictory. 
\item[](3) $\Pi_3=\{\, not\,p\leftarrow q,\;\;\leftarrow not\,p\,\}$, which is equivalent to $\{\,\leftarrow p,q,\;\; \leftarrow not\,p\,\}$ (Proposition~\ref{ic-trans}), is incoherent, but 
$WDA(\Pi_3)=\Pi_3\cup\{\, p\leftarrow not\,q\,\}$ is consistent (having the answer set $\{p\}$). 
\item[](4)  $\Pi_4=\{\, \leftarrow not\,p,\;\; \neg\, p\leftarrow not\,q,\;\; q\leftarrow\,\}$ is incoherent, but 
$SDA(\Pi_4)=\Pi_4\cup\{\, p\leftarrow q\,\}$ is consistent (having the answer set $\{p,q\}$). 
\end{description}
\end{example}


\section{AC and DA as Default Reasoning} \label{sec:4}

AC and DA are logically invalid and additional rules for AC and DA often make a program inconsistent. 
In this section, we relax the effects of the AC or DA completion by introducing additional rules as {\em default rules\/} 
in the sense of \cite{Rei80}.  
More precisely, we capture AC and DA as the following default inference rules: 
\begin{eqnarray*} 
({\bf default\, AC}) &&\frac{(\phi\Rightarrow \psi)\wedge \psi : \phi}{\phi} \\
({\bf default\, DA}) && \frac{(\phi\Rightarrow \psi)\wedge \neg\phi : \neg\psi}{\neg\psi} 
\end{eqnarray*}
The default AC rule says: given the conditional ``$\phi\Rightarrow \psi$" and the fact $\psi$, 
conclude $\phi$ as a default consequence.  
Likewise, the default DA rule says: given the conditional ``$\phi\Rightarrow \psi$" and the fact $\neg\phi$, 
conclude $\neg\psi$ as a default consequence. 
We encode these rules in ASP. 

\subsection{Default AC completion}

The AC completion is modified for default AC reasoning. 

\begin{definition}[default AC completion]\rm
Let $\Pi$ be a program. For each rule $r\in\Pi$ of the form: 
\begin{eqnarray*}
&& L_1\,;\,\cdots\,;\, L_k\, ;\, not\,L_{k+1}\,;\cdots ;\,not\,L_l\\
&& \qquad \leftarrow\; L_{l+1},\,\ldots,\,L_m,\,not\,L_{m+1},\,\ldots,\,not\,L_n, 
\end{eqnarray*}
define $dac(r)$ as the set of rules: 
\begin{eqnarray}
&& L_{l+1},\,\ldots,\,L_m,\, not\,L_{m+1},\,\ldots,\,not\,L_n \leftarrow L_i,\,\Delta\;\;\;\;\;\quad (1\leq i\leq k), \label{drev-1}\\
&& L_{l+1},\,\ldots,\,L_m,\, not\,L_{m+1},\,\ldots,\,not\,L_n \leftarrow not\,L_i,\,\Delta\;\;\;\; (k+1\leq i\leq l)\label{drev-2}
\end{eqnarray}
where $\Delta=not\,\neg\,L_{l+1},\,\ldots,\,not\,\neg\, L_m,\, not\,L_{m+1},\,\ldots,\,not\,L_n$. 
As before, (\ref{drev-1}) is not produced if $head^+(r)=\emptyset$ or $body^+(r)=body^-(r)=\emptyset$; and 
(\ref{drev-2}) is not produced if $head^-(r)=\emptyset$ or $body^+(r)=body^-(r)=\emptyset$. 
The {\em default AC completion\/} of $\Pi$ is defined as: 
\[ DAC(\Pi) = \Pi \;\cup\; dac(\Pi)\] 
in which 
\begin{eqnarray*} 
dac(\Pi)&\!=\!& \{\; \Sigma_1\,;\,\cdots\, ;\,\Sigma_p \;\leftarrow \; \ell_j,\,\Delta_i  \;\mid\\
&& \quad \Sigma_i\leftarrow \ell_j,\,\Delta_i \;\, (1\le i\le p)\;\;\mbox{is in}\;\; \bigcup_{r\in\Pi}\, dac(r)\, \} 
\end{eqnarray*}
where each $\Sigma_i$ $(1\le i\le p)$ is a conjunction of literals and NAF-literals,  
and $\ell_j$ is either a literal $L_j$ $(1\le j\le k)$ or an NAF-literal $not\,L_j$ $(k+1\le j\le l)$. 
\end{definition}

Like $AC(\Pi)$, rules in $dac(\Pi)$ are converted into the form of a GEDP, then $DAC(\Pi)$ is viewed as a GEDP. 
Compared with the AC completion, the DAC completion introduces the conjunction $\Delta$ of NAF-literals to the body of each rule.  Then the rules ``$\Sigma_1\,;\,\cdots\, ;\,\Sigma_p \,\leftarrow \, \ell_j,\,\Delta_i$" having the same head with 
different bodies are constructed for $i=1,\ldots,p$. 

\begin{example}
Let $\Pi=\{\,p\leftarrow q,\;\; p\leftarrow r,\;\; p\leftarrow,\;\; \neg r\leftarrow\,\}$. 
Then $DAC(\Pi)=\Pi\cup dac(\Pi)$ where 
\[ dac(\Pi)=\{\, q\,; r\leftarrow p,not\,\neg q,\;\;\; q\,; r\leftarrow p,not\,\neg r\,\}. \]
As a result, $DAC(\Pi)$ has the answer set $\{p,q,\neg r\}$. 
\end{example}

We say that a set $S$ of ground literals {\em satisfies\/} 
the conjunction ``$L_1,\ldots,L_k$" of ground literals if $\{L_1,\ldots,L_k\}\subseteq S$; and 
$S$ {\em satisfies\/} the conjunction ``$not\,L_1,\ldots,not\,L_k$" of ground NAF-literals if $\{L_1,\ldots,L_k\}\cap S=\emptyset$. 

When $AC(\Pi)$ has a consistent answer set, $DAC(\Pi)$ does not change it. 

\begin{proposition} \label{consistent-dac}\em
Let $\Pi$ be a program. If $AC(\Pi)$ has a consistent answer set $S$, then  
$S$ is an answer set of $DAC(\Pi)$. 
\begin{proof}\rm 
Suppose that $AC(\Pi)$ has a consistent answer set $S$. 
For each rule 
($\Sigma_1\,;\,\cdots\, ;\,\Sigma_p \;\leftarrow \; L_j$) $(1\le j\le k)$ in $ac(\Pi)$, 
$L_j\in S$ implies that $S$ satisfies some conjunction $\Sigma_i$ $(1\le i\le p)$. 
In this case, $S$ satisfies $\Delta_i$ and $L_j\in S$ implies that $S$ satisfies $\Sigma_i$ for each rule 
($\Sigma_1\,;\,\cdots\, ;\,\Sigma_p \;\leftarrow \; L_j,\Delta_i$) in $dac(\Pi)$. 
Likewise, for each rule 
($\Sigma_1\,;\,\cdots\, ;\,\Sigma_p \;\leftarrow \; not\,L_j$) $(k+1\le j\le l)$ in $ac(\Pi)$, 
$L_j\not\in S$ implies that $S$ satisfies some $\Sigma_i$ $(1\le i\le p)$. 
In this case, $S$ satisfies $\Delta_i$ and $L_j\not\in S$ implies that $S$ satisfies $\Sigma_i$ for each rule 
($\Sigma_1\,;\,\cdots\, ;\,\Sigma_p \;\leftarrow \; not\,L_j,\Delta_i$) in $dac(\Pi)$. 
Then $AC(\Pi)^S=DAC(\Pi)^S$ and the result follows. 
\end{proof}
\end{proposition}

DAC does not introduce the contradictory answer set $Lit$ unless the original program has $Lit$ as an answer set. 

\begin{proposition} \label{dac-prop}\em
Let $\Pi$ be a program.  If $DAC(\Pi)$ has the answer set $Lit$, then $\Pi$ has the answer set $Lit$. 
\begin{proof}\rm
If $DAC(\Pi)$ has the answer set $Lit$, then 
$DAC(\Pi)^{Lit}=\Pi^{Lit}\cup dac(\Pi)^{Lit}$ has the answer set $Lit$. 
Since $dac(\Pi)^{Lit}=\emptyset$, $\Pi^{Lit}$ has the answer set $Lit$. 
Hence, $\Pi$ has the answer set $Lit$. 
\end{proof}
\end{proposition}

$DAC(\Pi)$ possibly turns a contradictory $AC(\Pi)$ into a consistent program. 

\begin{example}[cont.\ Example~\ref{inconsistent-ac-ex}]\rm
Let $\Pi_1=\{\, p\leftarrow \neg\, p,\;\; p\leftarrow \,\}$. 
Then, $DAC(\Pi_1)=\Pi_1\cup \{\, \neg\, p\leftarrow p,\, not\,p \,\}$ has the single answer set $\{\,p\,\}$. 
So $AC(\Pi_1)$ is contradictory, but $DAC(\Pi_1)$ is consistent. 
\end{example}

When $AC(\Pi)$ is incoherent, $DAC(\Pi)$ does not resolve incoherency in general. 

\begin{example}\rm
Let $\Pi=\{\, p\leftarrow q,\;\; p\leftarrow,\;\; \leftarrow q \,\}$. 
Then, $AC(\Pi)=\Pi\cup \{\, q\leftarrow p \,\}$ is incoherent. 
$DAC(\Pi)=\Pi\cup \{\, q\leftarrow p,\, not\,\neg\,q \,\}$ is still incoherent. 
\end{example}

\subsection{Default DA completion}

The DA completion is modified for default DA reasoning. 

\begin{definition}[default DA completion]\rm
Let $\Pi$ be a program.  Define 
\begin{eqnarray*}
wdda(\Pi) &\!=\!& \{\, \ell_i \leftarrow \Gamma_1,\ldots, \Gamma_p,\,\delta^w_i \,\mid \\
&& \quad \ell_i\leftarrow \Gamma_j\; (1\leq j\leq p)\;\;\mbox{is in}\;\; \bigcup_{r\in\Pi}\; winv(r)\,\},\\
sdda(\Pi) &\!=\!& \{\, \ell_i \leftarrow \Gamma_1,\ldots, \Gamma_p,\,\delta^s_i \,\mid \\
&& \quad \ell_i\leftarrow \Gamma_j\; (1\leq j\leq p)\;\;\mbox{is in}\;\; \bigcup_{r\in\Pi}\; sinv(r)\,\}
\end{eqnarray*}
where $\ell_i$, $\Gamma_j$, $winv(r)$, and $sinv(r)$ 
are the same as those in Defs.~\ref{wda-def} and~\ref{sda-def}. 
In addition,  
$\delta^w_i=not\,\neg\,L_i$ if $\ell_i=L_i$, and $\delta^w_i=not\,L_i$ if $\ell_i=not\,L_i$; 
$\delta^s_i=not\,\neg\,L_i$ if $\ell_i=L_i$, and $\delta^s_i=not\,L_i$ if $\ell_i=\neg\,L_i$. 
The {\em weak default DA completion\/} and the {\em strong default DA completion\/} 
of $\Pi$ are respectively defined as:
\begin{eqnarray*}
WDDA(\Pi) &=& \Pi \;\cup\; wdda(\Pi),\\ 
SDDA(\Pi) &=& \Pi \;\cup\; sdda(\Pi). 
\end{eqnarray*}
\end{definition}

Rules in $wdda(\Pi)$ and $sdda(\Pi)$ are converted into the form of a GEDP, so $WDDA(\Pi)$ and $SDDA(\Pi)$ are viewed as GEDPs. 
Like the DAC completion, both WDDA and SDDA introduce an additional NAF-literal to each rule. 

When $WDA(\Pi)$ (resp.\ $SDA(\Pi)$) has a consistent answer set, $WDDA(\Pi)$ (resp.\ $SDDA(\Pi)$) does not change it. 

\begin{proposition} \em
Let $\Pi$ be a program. If\, $WDA(\Pi)$ (resp.\ $SDA(\Pi)$) has a consistent answer set $S$, then  
$S$ is an answer set of $WDDA(\Pi)$ (resp.\ $SDDA(\Pi)$). 
\begin{proof}\rm 
Suppose that $WDA(\Pi)$ has a consistent answer set $S$. 
If $L_i\in S$ then $\neg\,L_i\not\in S$ and $\delta^w_i=not\,\neg\,L_i$ is eliminated in $WDDA(\Pi)^S$. 
Else if $L_i\not\in S$ and $\neg\,L_i\in S$, there is another rule 
($\ell_i \leftarrow \Gamma_1,\ldots, \Gamma_p,\,\delta^w_i$) in $wdda(\Pi)$ such that $\ell_i=\neg\,L_i$ and $\delta^w_i=not\,L_i$. 
Then, $\delta^w_i=not\,L_i$ is eliminated in $WDDA(\Pi)^S$. 
Thus, $WDA(\Pi)^S=WDDA(\Pi)^S$. 
Similarly, $SDA(\Pi)^S=SDDA(\Pi)^S$. 
Hence, the result follows. 
\end{proof}
\end{proposition}

WDDA or SDDA does not introduce the contradictory answer set $Lit$ unless the original program has $Lit$ as an answer set. 

\begin{proposition}\em \label{consistent-dda-prop}
Let $\Pi$ be a program.  If $WDDA(\Pi)$ (or $SDDA(\Pi)$) has the answer set $Lit$, then $\Pi$ has the answer set $Lit$. 
\begin{proof}\rm 
If $WDDA(\Pi)$ has the answer $Lit$, 
then $WDDA(\Pi)^{Lit}=\Pi^{Lit}\cup wdda(\Pi)^{Lit}$ has the answer set $Lit$. 
Since $wdda(\Pi)^{Lit}=\emptyset$, $\Pi^{Lit}$ has the answer set $Lit$. 
Hence, $\Pi$ has the answer set $Lit$. 
The case of $SDDA(\Pi)$ is proved in a similar manner. 
\end{proof}
\end{proposition}

$WDDA(\Pi)$  (resp.\ $SDDA(\Pi)$) possibly turns a contradictory $WDA(\Pi)$  (resp.\ $SDA(\Pi)$) into a consistent program, while it 
does not resolve incoherency in general. 

\begin{example}[cont.\ Example~\ref{inconsistent-da-ex}]\rm
Let $\Pi_1=\{\, not\,p\leftarrow p \,\}$ where $WDA(\Pi_1)$ is incoherent. 
$WDDA(\Pi_1)=\Pi_1\cup \{\, p\leftarrow not\,p,\, not\,\neg\,p \,\}$ is still incoherent. 
Let $\Pi_2=\{\, \neg\, p\leftarrow p,\;\;\; \neg\, p\leftarrow \,\}$ where $SDA(\Pi_2)$ is contradictory. 
$SDDA(\Pi_2)=\Pi_2\cup \{\, p\leftarrow \neg\,p,\, not\,\neg \,p \,\}$ has the consistent answer set $\{\neg p\}$. 
\end{example}

\section{Characterizing Human Reasoning Tasks} \label{sec:5}

\subsection{Suppression Task} \label{sec:5.1}
 
\cite{Byr89} provides an empirical study which shows that human conditional reasoning can be nonmonotonic 
in the so-called {\em suppression tasks}. 
She verifies the effects in different types of conditional reasoning by experimental testing on college students. 
Students are divided into three groups: 
the first group receives simple conditional arguments; 
the second group receives conditional arguments accompanied by another conditional sentence with an alternative antecedent; 
and the third group receives conditional arguments accompanied by another conditional sentence with an additional antecedent. 
More precisely, suppose the following three conditional sentences: 
\begin{quote}
$S_1$:\, If she has an essay to write then she will study late in the library. 

$S_2$:\, If she has some textbooks to read then she will study late in the library. 

$S_3$:\, If the library stays open then she will study late in the library.
\end{quote}
Given the conditional sentence $S_1$,  $S_2$ represents a sentence with an {\em alternative\/} antecedent, while 
$S_3$ represents a sentence with an {\em additional\/} antecedent. 
The antecedent of $S_2$ is considered an alternative sufficient condition for studying late in the library. 
By contrast, the antecedent of $S_3$ is considered an additional sufficient condition for studying late in the library. 
Table~\ref{table-byrne} presents the percentages of inferences made by subjects from the three kinds of arguments.

\begin{table}[t]
\caption{The percentages of inferences in experiments (\cite{Byr89}) } \label{table-byrne}
\centering
\begin{tabular}{ccccc}\hline
Argument Type\,\,\,\,\,\, &  \,\,\,AA (MP)\,\,\, & \,\,\,DC (MT)\,\,\, & \,\,\,\,AC\,\,\,\,\,\, & \,\,\,\,\,\,\,DA\,\,\,\, \\ \hline
simple arguments ($S_1$) & 96\% & 92\% & 71\% & 46\% \\
alternative arguments ($S_1$ and $S_2$) & 96\% & 96\% & 13\% & 4\% \\
additional arguments ($S_1$ and $S_3$) & 38\% & 33\% & 54\% & 63\% \\
\hline
\end{tabular}
\end{table}

By the table, given the sentence $S_1$ and the fact: ``she will study late in the library", 
the 71\% of the first group concludes: ``she has an essay to write" by AC. 
When $S_1$ is accompanied by a conditional $S_2$ containing an alternative antecedent, on the other hand, 
the percentage of subjects who perform AC inference reduces to 13\%. 
The reason is that people know that the alternative: ``She has some textbooks to read'', could be the case instead. 
Similar reduction is observed for DA. 
Byrne argues that those fallacious inferences are {\em suppressed\/} when a conditional is 
accompanied by an alternative antecedent. 
She also observes that the inference patterns change 
when $S_1$ is accompanied by a conditional $S_3$  containing an additional antecedent. 
In Table~\ref{table-byrne} 
the number of subjects who conclude: ``She will study late in the library" by AA reduces to 38\%, 
and the number of subjects who conclude: ``She does not have an essay to write" by DC reduces to 33\%. 
By contrast, the suppression of AC and DA are relaxed, 54\% of subjects make AC and 63\% of subjects make DA. 
Byrne then argues that ``valid inferences are suppressed in the same way as fallacious inferences". 

The suppression task is characterized in our framework as follows. 
First, the sentence $S_1$ is represented as the rule: ``$library \leftarrow essay$". 
Then, four conditional inferences (AA, DC, AC, DA) in simple arguments are respectively represented by the following programs: 
\begin{eqnarray*}
\mbox{(AA)} && \Pi_0=\{\,  library \leftarrow essay,\;\;\; essay\leftarrow\,\}, \\
\mbox{(DC)} && \Pi_1=\{\,  library \leftarrow essay,\;\;\; \neg\, library\leftarrow\,\}, \\
\mbox{(AC)} && \Pi_2=\{\,  library \leftarrow essay,\;\;\; library\leftarrow\,\}, \\
\mbox{(DA)} && \Pi_3=\{\,  library \leftarrow essay,\;\;\; \neg\,essay\leftarrow\,\}. 
\end{eqnarray*}
Then $\Pi_0$ has the answer set $\{\,library, essay\,\}$ in which AA inference is done. 
By contrast, DC, AC, and DA inferences are not performed in $\Pi_1$, $\Pi_2$, and $\Pi_3$, respectively. 
To realize those inferences, consider completions such that 
\begin{eqnarray*}
SDC(\Pi_1)&=&\Pi_1\cup \{\, \neg\,essay \leftarrow \neg\,library\,\}, \\
AC(\Pi_2)&=&\Pi_2\cup\{\,  essay\leftarrow library \,\}, \\
SDA(\Pi_3)&=&\Pi_3\cup\{\,  \neg\,library \leftarrow \neg\,essay\,\}
\end{eqnarray*}
where $SDC(\Pi_1)$ has the answer set $\{\,\neg\,library, \neg\,essay\,\}$, 
$AC(\Pi_2)$ has the answer set $\{\,library, essay\,\}$, and 
$SDA(\Pi_3)$ has the answer set $\{\,\neg\,library, \neg\,essay\,\}$. 
As a result, DC, AC, and DA inferences are performed in $SDC(\Pi_1)$, $AC(\Pi_2)$, and $SDA(\Pi_3)$, respectively. 

Next, consider the alternative arguments $S_1$ and $S_2$.  
They are represented by the programs: \[\Pi_k^{\rm ALT}=\Pi_k \cup \{\, library\leftarrow text\,\}\;\;\; (k=0,1,2,3). \]  
The program $\Pi_0^{\rm ALT}$ has the answer set $\{\,library, essay\,\}$ in which the result of AA inference does not change from $\Pi_0$. 
Programs $\Pi_1^{\rm ALT}$, $\Pi_2^{\rm ALT}$, and $\Pi_3^{\rm ALT}$ are completed as follows: 
\begin{eqnarray*}
SDC(\Pi_1^{\rm ALT})&=&\Pi_1^{\rm ALT}\cup \{\, \neg\,essay \leftarrow \neg\,library,\;\;\; \neg\,text \leftarrow \neg\,library\,\}, \\
AC(\Pi_2^{\rm ALT})&=&\Pi_2^{\rm ALT}\cup\{\,  essay\,;\, text\leftarrow library \,\}, \\
SDA(\Pi_3^{\rm ALT})&=&\Pi_3^{\rm ALT}\cup\{\,  \neg\,library \leftarrow \neg\,essay,\,\neg\,text\,\}
\end{eqnarray*}
where $SDC(\Pi_1^{\rm ALT})$ has the answer set $\{\,\neg\,library, \neg\,essay, \neg\,text\,\}$, 
$AC(\Pi_2^{\rm ALT})$ has the two answer sets $\{\,library, essay\,\}$ and $\{\,library, text\,\}$, and 
$SDA(\Pi_3^{\rm ALT})$ has the answer set $\{\,\neg\,essay\,\}$. 
As a result, $SDC(\Pi_1^{\rm ALT})\models_s \neg\,essay$, $AC(\Pi_2^{\rm ALT})\not\models_s essay$, and 
$SDA(\Pi_3^{\rm ALT})\not\models_s \neg\,library$, which indicate that AC and DA inferences are suppressed while DC is not suppressed. 
In this way, the completion successfully represents the effect of suppression of AC/DA inference in alternative arguments. 

In additional arguments, on the other hand, \cite{Byr89} observes that AA/DC inference is also suppressed. 
Our completion method does not characterize the suppression of AA inference 
because we enable fallacious inferences by AC/DA completion while still keep the valid AA inference. 
\cite{Byr89} says ``{\em people may consider that certain other conditions are necessary
for this conclusion to hold, for example, the library must remain open. Thus,
conditionals are frequently elliptical in that information that can be taken for granted is omitted from them}". 
In the above example, the availability of the library is necessary for studying in it but it is just omitted in the initial premises. 
Then it is considered that the rule: ``$library\leftarrow essay$" 
in mind is overwritten by the rule: ``$library\leftarrow essay,\, open$" when the additional antecedent is given.  Let 
\begin{eqnarray*}
&& \Pi_0^{\rm ADD}=\{\,  library\leftarrow essay,\,open,\;\;\; essay\leftarrow\,\}, \\
&& \Pi_1^{\rm ADD}=\{\,  library\leftarrow essay,\,open,\;\;\; \neg\, library\leftarrow\,\}, \\
&& \Pi_2^{\rm ADD}=\{\,  library\leftarrow essay,\,open,\;\;\; library\leftarrow\,\}, \\
&& \Pi_3^{\rm ADD}=\{\, library\leftarrow essay,\,open,\;\;\; \neg\,essay\leftarrow\,\}.
\end{eqnarray*}
The program $\Pi_0^{\rm ADD}$ has the answer set $\{\,essay\,\}$, thereby $\Pi_0^{\rm ADD}\not\models_s library$. 
Then the result of AA inference is suppressed. 
Programs $\Pi_1^{\rm ADD}$, $\Pi_2^{\rm ADD}$, $\Pi_3^{\rm ADD}$ are completed as follows: 
\begin{eqnarray*}
SDC(\Pi_1^{\rm ADD})&=&\Pi_1^{\rm ADD}\cup \{\, \neg\,essay\,;\, \neg\,open \leftarrow \neg\,library\,\}, \\
AC(\Pi_2^{\rm ADD})&=&\Pi_2^{\rm ADD}\cup\{\,  essay\leftarrow library,\;\;\;  open\leftarrow library \,\}, \\
SDA(\Pi_3^{\rm ADD})&=&\Pi_3^{\rm ADD}\cup\;\{\,  \neg\,library \leftarrow \neg\,essay,\;\;\; \neg\,library \leftarrow \neg\,open\,\}
\end{eqnarray*}
where $SDC(\Pi_1^{\rm ADD})$ has the two answer sets $\{\,\neg\,library, \neg\,essay\,\}$ and $\{\,\neg\,library, \neg\,open\,\}$, 
$AC(\Pi_2^{\rm ADD})$ has the answer set $\{\,library, essay, open\,\}$, and 
$SDA(\Pi_3^{\rm ADD})$ has the the answer set $\{\,\neg\,essay, \neg\,library\,\}$. 
As a result, $SDC(\Pi_1^{\rm ADD})\not\models_s \neg\,essay$, $AC(\Pi_2^{\rm ADD})\models_s essay$, and 
$SDA(\Pi_3^{\rm ADD})\models_s \neg\,library$. This indicates that DC is suppressed but AC and DA are not suppressed, which 
explains the results of \cite{Byr89}.  

The results of inferences using completion are summarized in Table~\ref{table-comp}. 
By the table, the suppression of AC and DA in face of an alternative antecedent is realized in our framework. 
The suppression of AA and DC in face of an additional antecedent is also realized if the additional condition is written in the antecedent of the original conditional sentence. 

\begin{table}[t]
\caption{Summary of inferences made by completion} \label{table-comp}
\centering
\begin{tabular}{ccccc}\hline
Argument Type\,\,\,\,\,\, &  \,\,\,AA (MP)\,\,\, & \,\,\,DC (MT)\,\,\, & \,\,\,\,AC\,\,\,\,\,\, & \,\,\,\,\,\,\,DA\,\,\,\, \\ \hline
$S_1$ & $\circ$ ($\Pi_0$) &  $\circ$ (SDC($\Pi_1$)) & $\circ$ (AC($\Pi_2$)) & $\circ$ (SDA($\Pi_3$))\\
$S_1$ and $S_2$ &  $\circ$  ($\Pi_0^{\rm ALT}$) &  $\circ$ (SDC($\Pi_1^{\rm ALT}$)) & $\times$ (AC($\Pi_2^{\rm ALT}$))&  $\times$ (SDA($\Pi_3^{\rm ALT}$))\\
$S_1$ and $S_3$ & ($\times$ ($\Pi_0^{\rm ADD}$)) & ($\times$ (SDC($\Pi_1^{\rm ADD}$))) &   ($\circ$ (AC($\Pi_2^{\rm ADD}$))) &  ($\circ$ (SDA($\Pi_3^{\rm ADD}$))) \\
\hline
\multicolumn{5}{c}\mbox{$\circ$ means that inference succeeds; $\times$ means that inference is suppressed.}
\end{tabular}
\end{table}


\subsection{Wason Selection Task} \label{sec:5.2}

\cite{Was68} introduces the {\em selection task\/} for examining human conditional reasoning. 
The task is described as follows. 
There are four cards on a table, each of which has a letter on one side and a number on the other. 
Suppose that those cards are showing respectively $D$, $K$, $3$, $7$. 
Given the sentence: ``Every card which has the letter $D$ on one side has the number $3$ on the other side", then 
which cards are to be turned over in order to verify the truth of the sentence? 
Testing on college students, it turns out that a relatively small number of students select the logically 
correct answer `$D$ and $7$' (4\%), while others select `$D$ and $3$' (46\%) or $D$ alone (33\%) (\cite{WS71}). 
The result shows that people are likely to perform AC inference but less likely to perform logically correct DC inference in this task.  The situation is characterized in our framework as follows. 

The sentence: ``Every card which has the letter $D$ on one side has the number $3$ on the other side" is rephrased as 
``If a card has the letter $D$ on one side, then it has the number $3$ on the other side".  Then it is represented by the program: 
\begin{equation}\label{wason-rule}
\Pi_W =\{\, n_3 \leftarrow \ell_D \} 
\end{equation}
where $n_3$ means the number 3 and $\ell_D$ means the letter $D$. 
Four cards on the desk are represented by the facts: 
\begin{equation}\label{wason-fact}
 \ell_D\leftarrow,\quad \ell_K\leftarrow,\quad n_3\leftarrow,\quad n_7\leftarrow. 
\end{equation}
Then each card is checked one by one.  
\begin{itemize}
\item $\Pi_W\cup \{\ell_D\leftarrow\}$ has the answer set $\{\ell_D, n_3\}$. 
If the other side of $\ell_D$ is not the number 3, however, $\{\ell_D, n_3\}\cup\{\neg n_3\}$ is contradictory. 
To verify the consistency, one has to turn over the card of $D$. 
\item $\Pi_W\cup \{\ell_K\leftarrow\}$ has the answer set $\{\ell_K\}$. 
Since both $\{\ell_K\}\cup\{n_3\}$ and $\{\ell_K\}\cup\{\neg n_3\}$ are consistent, there is no need to turn over the card of $K$. 
\item $\Pi_W\cup \{n_3\leftarrow\}$ has the answer set $\{n_3\}$. 
Since both $\{n_3\}\cup\{\ell_D\}$ and $\{n_3\}\cup\{\neg \ell_D\}$ are consistent, there is no need to turn over the card of $3$. 
\item $\Pi_W\cup \{n_7\leftarrow\}$ has the answer set $\{n_7\}$. 
Since both $\{n_7\}\cup\{\ell_D\}$ and $\{n_7\}\cup\{\neg\ell_D\}$ are consistent, there is no need to turn over the card of $D$. 
\end{itemize}
As the standard ASP does not realize DC inference, it characterizes reasoners who select only $D$ as shown above. 

By contrast, people who choose $D$ and $3$ are likely to perform AC inference using the conditional sentence. 
In this case, the situation is represented as 
\[  AC(\Pi_W) = \{\, n_3 \leftarrow \ell_D,\;\;\; \ell_D\leftarrow n_3\,\}. \]
Now $AC(\Pi_W)\cup \{n_3\leftarrow\}$ has the answer set $\{n_3,\ell_D\}$. 
If the other side of $n_3$ is not the letter $D$, however, $\{n_3,\ell_D\}\cup\{\neg\ell_D\}$ is contradictory. 
To verify the consistency, they opt to turn over the card of $3$ as well as the card of $D$. 

Finally, those who choose $D$ and $7$ perform weak DC inference as%
\footnote{Alternatively, the SDC ``$\neg\,\ell_D\leftarrow\neg\, n_3$" is used by introducing 
the rule ``$\neg\,n_3\leftarrow n_7$" instead of the constraint ``$\leftarrow n_3, n_7$".}
\[ 
 \mathit{WDC}(\Pi_W)=\{\, n_3 \leftarrow \ell_D,\;\;\; not\,\ell_D\leftarrow not\,n_3\,\}.\]
The program $\mathit{WDC}(\Pi_W)\cup \{n_7\leftarrow\}$ has the answer set $\{n_7\}$. 
However, if the other side of $n_7$ is the letter $D$, 
$\mathit{WDC}(\Pi_W)\cup \{n_7\leftarrow\}\cup\{ \ell_D\leftarrow\}\cup \{\leftarrow n_3,n_7\}$ is incoherent, 
where the constraint ``$\leftarrow n_3, n_7$" represents that one card cannot have two numbers. 
To verify the consistency, they need to turn over the card of 7 as well as the card of $D$.%
\footnote{\cite{Kow11} also uses an integrity constraint to explain the effect of modus tollens in the selection task. }

It is known that the Wason selection task is context dependent and the results change when, for instance, it is presented with a deontic rule. 
\cite{GC82} use the rule: ``If a person is drinking beer, then the person must be over 19 year of age". 
There are four cards on a table as before, but this time a person's age is on one side of a card and on the other side is what the 
person is drinking.  Four cards show, `beer', `coke', `16', and `22', respectively. 
Then select the card(s) needed to turn over to determine whether or not no person is violating the rule. 
In this drinking-age problem, almost 75\% of participants select logically correct answers `beer' and `16'. 
To characterize the situation, (W)DC completion is used for representing selection tasks in deontic contexts. 

\section{Applications to Commonsense Reasoning} \label{sec:6}

The AC, DC, DA completions realize human conditional reasoning in ASP. 
In addition, they are used for computing commonsense reasoning in AI. 

\subsection{Abduction and Prediction} \label{sec:6.1}

{\em Abduction\/} reasons from an observation to explanations. 
An {\em abductive logic program\/} (\cite{KKT92}) is defined as a pair $\pf{\Pi,\Gamma}$ where $\Pi$ is a program and 
$\Gamma\, (\subseteq Lit$) is a set of literals called {\em abducibles}.%
\footnote{\cite{KKT92} consider {\em integrity constraints\/} which are handled as constraints in $\Pi$.}
It is assumed that abducibles appear in the head of no rule in $\Pi$. 
Given an {\em observation\/} $O$ as a ground literal, the abduction problem is to find an {\em explanation\/} 
$E\,(\subseteq \Gamma)$ satisfying (i) $\Pi\cup E\models_x O$ and (ii) $\Pi\cup E$ is consistent, 
where $\models_x$ is either $\models_c$ or $\models_s$ depending on the problem.  
Here we consider $\models_c$ that realizes {\em credulous\/} abduction. 
In GEDP, the abduction problem is characterized as follows.  Let us define 
\[abd(\Gamma)=\{\,  \gamma\,;\,not\,\gamma\,\leftarrow\;\mid\; \gamma\in\Gamma\,\}.\] 

\begin{proposition}[\cite{IS98}]\label{prop-is98}\em
Let $\pf{\Pi,\Gamma}$ be an abductive program.  Given an observation $O$, a set $E\subseteq\Gamma$ is an 
explanation of $O$ iff\, $\Pi\cup abd(\Gamma)\cup \{\,\leftarrow not\,O \,\}$ has a consistent answer set $S$ such that 
$S\cap\Gamma=E$. 
\end{proposition}

\begin{example}\rm \label{abduction-ex}
Consider $(\Pi_1,\Gamma_1)$ where  
\begin{eqnarray*}
\Pi_1:&& arrive\_on\_time\leftarrow not\;accident,\\
&& \neg\,arrive\_on\_time\leftarrow accident. \\
\Gamma_1:&& accident. 
\end{eqnarray*}
$\Pi_1$ represents that a train arrives on time unless there is an accident. 
Given the observation $O=\neg\,arrive\_on\_time$, it has the explanation $E=\{accident\}$. 
The problem is represented as the GEDP: 
\begin{eqnarray*}
&& arrive\_on\_time\leftarrow not\;accident,\\
&& \neg\,arrive\_on\_time\leftarrow accident, \\
&& accident\,;\, not\,accident\leftarrow,\\
&& \leftarrow not\,\neg\,arrive\_on\_time
\end{eqnarray*}
which has the answer set $S=\{ \neg\,arrive\_on\_time,\; accident \}$ 
where $S\cap\Gamma_1=E$. 
\end{example}

Since abduction reasons backward from an observation, it is characterized using AC inference as follows. 

\begin{proposition}\em \label{ac-abd-prop}
Let $\pf{\Pi,\Gamma}$ be an abductive program and $O$ an observation. 
\begin{description}
\item[](i) A set $E\subseteq\Gamma$ is an explanation of $O$ if $O\in head^+(r)$ for some $r\in\Pi$ and 
$AC(\Pi)\cup \{ O\}$ has a consistent answer set $S$ such that $S\cap\Gamma=E$. 
\item[](ii) If a set $E\subseteq\Gamma$ is an explanation of $O$, then there is $\Pi'\subseteq\Pi$ such that 
$AC(\Pi')\cup \{ O\}$ has a consistent answer set $S$ such that $S\cap\Gamma=E$. 
\end{description}
\begin{proof}\rm
(i) Suppose that $O\in head^+(r)$ for some $r\in\Pi$ and $AC(\Pi)\cup \{ O\}$ has a consistent answer set $S$ such that $S\cap\Gamma=E$.  
By the definition of $AC(\Pi)$, there is a path from $O$ to 
literals in $E$ in the dependency graph\footnote{A not-free EDP $\Pi$ is associated with a {\em dependency graph\/} $(V,E)$ 
where the nodes $V$ are literals of $\Pi$ and there are edges in $E$ from $L\in  body^+(r)$ 
to $L'\in head^+(r)$ for each $r\in\Pi$. }  
of the program $AC(\Pi)^S\cup\{O\}$ where $AC(\Pi)^S$ is the reduct of $AC(\Pi)$ by $S$.  
In this case, $O$ is reached by reasoning forward from $E$ in $\Pi^S$, and 
$S$ is an answer set of $\Pi^S\cup  E$ such that $O\in S$. 
This implies that $\Pi\cup  E\cup\{\leftarrow not\,O\}$ has a consistent answer set $S$ such that $abd(\Gamma)^S=E$, and 
the result holds by Proposition~\ref{prop-is98}  

(ii) If $E\subseteq\Gamma$ is an explanation of $O$, there is $\Pi'\subseteq\Pi$ such that 
$\Pi'\cup abd(\Gamma)\cup\{\leftarrow not\,O\}$ 
has a consistent answer set $S$ satisfying $S\cap\Gamma=E$ (Proposition~\ref{prop-is98}). 
Select a minimal set $\Pi'$ of rules such that there is no $\Pi''\subset\Pi'$ satisfying the above condition. 
In this case, $abd(\Gamma)^S=E$ is obtained by reasoning backward from $O$ in $(\Pi')^S$, and 
$S$ is a minimal set satisfying $(\Pi')^S\cup ac(\Pi')^S\cup\{O\}$. 
Then $AC(\Pi')\cup \{ O\}$ has a consistent answer set $S$ such that $S\cap\Gamma=E$. 
\end{proof}
\end{proposition}

In Example~\ref{abduction-ex}, $AC(\Pi_1)\cup \{ O\}$ becomes 
\begin{eqnarray*}
&& arrive\_on\_time\leftarrow not\;accident,\\
&& \neg\,arrive\_on\_time\leftarrow accident. \\
&& not\;accident\leftarrow arrive\_on\_time,\\
&& accident\leftarrow \neg\,arrive\_on\_time,\\
&& \neg\,arrive\_on\_time \leftarrow, 
\end{eqnarray*}
which has the answer set $S=\{\, \neg\,arrive\_on\_time,\, accident\,\}$.  
By $S\cap\Gamma_1=\{\,accident\,\}$, $E=\{\,accident\,\}$ is the explanation. 

Note that 
$AC(\Pi)$ introduces converse of every rule, while explanations are computed using the AC completion of a subset $\Pi'\subseteq \Pi$ 
in general (Proposition~\ref{ac-abd-prop}(ii)). 

\begin{example}\rm \label{ex-abd-conv}
Let $\Pi=\{\, p\leftarrow a,\;\; q\leftarrow \neg a,\;\; q\leftarrow\,\}$ and $\Gamma=\{a,\neg a\}$. 
Then $O=p$ has the explanation $E=\{a\}$ in $\pf{\Pi,\Gamma}$, while 
$AC(\Pi)\cup\{O\}=\Pi\cup\{\, a\leftarrow p,\;\; \neg a\leftarrow q,\;\; p\leftarrow\,\}$ is contradictory. 
By putting $\Pi'=\{\, p\leftarrow a\,\}$, $AC(\Pi')\cup\{O\}$ has the consistent answer set $S=\{p,a\}$ where $S\cap\Gamma=\{a\}$. 
\end{example}
As illustrated in Example~\ref{ex-abd-conv}, abduction and AC completion produce different results in general. 

Abductive logic programs of (\cite{KKT92}) cannot compute explanations when contrary to the consequent is observed. 
For instance, consider $\pf{\Pi_2,\Gamma_2}$ such that 
\begin{eqnarray*}
\Pi_2:&& arrive\_on\_time\leftarrow not\;accident.\\
\Gamma_2:&& accident. 
\end{eqnarray*}
Given the observation $O=\neg\,arrive\_on\_time$, 
no explanation is obtained from $\pf{\Pi_2,\Gamma_2}$. 
Generally, a program $\Pi$ does not necessarily contain a pair of rules $r$ and $r'$ that define  
$L$ and $\neg L$, respectively.   When there is a rule defining $L$ but no rule defining $\neg L$, abduction computes 
no explanation for the observation $O=\neg L$.  The problem is resolved by reasoning by DC. 
For the rule $r$ in $\Pi_2$, $sdc(r)$ becomes 
\[  accident\leftarrow \neg\, arrive\_on\_time.\]
Then $SDC(\Pi_2)\cup\{O\}$ computes the explanation $\{\,accident\,\}$.  

In contrast to $SDC$, $WDC$ is used for abduction from {\em negative\/} observations. 
A negative observation represents that some evidence $G$ is not observed and it is represented as 
$O=not\,G$, 
which should be distinguished from the (positive) observation $O=\neg\,G$ meaning that $\neg\,G$ is observed. 
In the abductive program $\pf{\Pi_2,\Gamma_2}$, the negative observation $O=not\,arrive\_on\_time$ 
is explained using $wdc(r)$:  
\[  accident\leftarrow not\; arrive\_on\_time.\]
Then $WDC(\Pi_2)\cup\{O\}$ has the answer set $\{\,accident\,\}$. 

In this way, both AC and DC are used for computing explanations deductively, and 
DC is used for computing explanations that are not obtained using the framework of (\cite{KKT92}). 
Moreover, AC and DC realize {\em prediction\/} by combining abduction and deduction. 
\vspace{1cm}

\begin{example}\rm \label{prediction-ex}
Consider $(\Pi_3,\Gamma_3)$ where 
\begin{eqnarray*}
\Pi_3:&& arrive\_on\_time\leftarrow not\;accident,\\
&& \neg\,arrive\_on\_time\leftarrow accident, \\
&& newspaper \leftarrow accident.\\
\Gamma_3:&& accident. 
\end{eqnarray*}
The third rule in $\Pi_3$ says: if there is an accident,  a newspaper will report it. 
Given the observation $O=\neg\,arrive\_on\_time$, 
$AC(\Pi_3)\cup \{ O\}\models_s newspaper$. 
\end{example}

As such, AC or DC combines abduction and deduction to realize prediction. 

\subsection{Counterfactual Reasoning} \label{sec:6.2}

A {\em counterfactual\/} is a conditional statement representing what would be the case if its premise were true 
(although it is not true in fact). 
\cite{Lew73} introduces two different types of counterfactual sentences. 
Given two different events $\phi$ and $\psi$, two counterfactual sentences are considered: 
``if it were the case that $\phi$, then it would be the case that $\psi$"  
(written $\phi\,\Box\!\!\!\rightarrow\,\psi$) and ``if it were the case that $\phi$, then it might be the case that $\psi$" (written $\phi\,\Diamond\!\!\!\rightarrow\,\psi$). Here $(\phi\,\Box\!\!\!\rightarrow\,\psi)$ implies $(\phi\,\Diamond\!\!\!\rightarrow\,\psi)$. 
We consider counterfactual reasoning such that what would be the case if some facts were not true. 
We then realize Lewis's two types of counterfactual reasoning using DA inference in ASP. 

\begin{definition}[counterfact]\rm 
Let $\phi$ be a fact of the form: 
\[ L_1\,;\,\cdots\,;\, L_k\, ;\, not\,L_{k+1}\,;\cdots ;\,not\,L_l\leftarrow \quad (1\le k\le l).\] 
Then the {\em counterfact\/} $\overline{\phi}$ is the set of facts such that 
\[ \overline{\phi}=\{\,\neg\,L_i\leftarrow \,\mid\, 1\le i\le k\,\}\;\cup\; \{\, L_j\leftarrow\,\mid\, k+1\le j\le l\,\}.\]
Given a set $\Sigma$ of facts, 
define \[\overline{\Sigma}=\bigcup_{\phi\in\Sigma} \overline{\phi}.\] 
\end{definition}

We say that $\overline{\Sigma}$ {\em satisfies\/} the conjunction 
``$L_{l+1},\,\ldots,\,L_m,\,not\,L_{m+1},\,\ldots,\,not\,L_n$" if 
$\{\, L_i\leftarrow \,\mid\, l+1\le i\le m\}\subseteq\overline{\Sigma}$ and 
$\{\, L_j\leftarrow \,\mid\, m+1\le i\le n\}\cap\overline{\Sigma}=\emptyset$. 

\begin{definition}[counterfactual program]\rm \label{def-cf}
Let $\Pi$ be a program and $\Sigma$ a set of facts in $\Pi$.  Then, a {\em counterfactual program\/} $\Omega$ is defined as 
\[
\Omega=(\Pi\setminus\Sigma)\;\cup\; \overline{\Sigma} \;\cup\; sda(\Pi). 
\]
For $\lambda\in Lit$, define 
\begin{eqnarray*}
&& \overline{\Sigma}\;\Box\!\!\!\rightarrow\,\lambda\;\;\mbox{if}\;\; \Omega\models_s\lambda,\\
&& \overline{\Sigma}\;\Diamond\!\!\!\rightarrow\,\lambda\;\;\mbox{if}\;\;\Omega\models_c\lambda.
\end{eqnarray*}
\end{definition}

By definition,  $\Omega$ is obtained from $\Pi$ by removing a set $\Sigma$ of facts, and instead introducing a set 
$\overline{\Sigma}$ of counterfacts as well as strong DA rules $sda(\Pi)$. 
$\overline{\Sigma}\;\Box\!\!\!\rightarrow\,\lambda$ (resp.\ $\overline{\Sigma}\;\Diamond\!\!\!\rightarrow\,\lambda$) 
means that 
if the counterfacts $\overline{\Sigma}$ were the case, then $\lambda$ is included in every (resp.\ some) answer set of $\Omega$. 

\begin{example}\rm 
Consider the program $\Pi$: 
\begin{eqnarray*}
&& London\,;\,Paris \leftarrow not\; virtual,\\
&& virtual\leftarrow pandemic,\\
&& pandemic \leftarrow. 
\end{eqnarray*}
An event is scheduled to take place in either London or Paris if it is not virtual. 
The pandemic turns the event into virtual, however. 
Suppose a counterfactual sentence: ``{\em if there were no pandemic, the event would not be virtual}". 
Putting $\Sigma=\{\,pandemic\,\}$,  
$\Omega$ becomes: 
\begin{eqnarray*}
&& London\,;\,Paris \leftarrow not\; virtual,\\
&& virtual\leftarrow pandemic,\\
&& \neg\,London\leftarrow virtual,\\
&& \neg\,Paris\leftarrow virtual,\\
&& \neg\,virtual\leftarrow \neg\,pandemic,\\
&& \neg\,pandemic \leftarrow. 
\end{eqnarray*}
Then, $\Omega$ has two answer sets:  
\begin{eqnarray*}
&& \{\,  \neg\,pandemic,\; \neg\,virtual,\; London\,\},\\
&& \{\, \neg\,pandemic,\; \neg\,virtual,\; Paris\,\}.
\end{eqnarray*} 
As a result, it holds that 
\[\{\neg\,pandemic\}\,\Box\!\!\!\rightarrow\,\neg\,virtual\quad\mbox{and}\quad  
\{\neg\,pandemic\}\,\Diamond\!\!\!\rightarrow\,London. \] 
\end{example}

Given a consistent program $\Pi$, it might be the case that the program $\Omega$ is inconsistent. 
To eliminate contradictory $\Omega$, default DA completion is used instead of DA completion. 
Define $\Omega_D$ that is obtained by replacing $sda(\Pi)$ by its default version $sdda(\Pi)$ in Definition~\ref{def-cf}. 
The next result holds by Proposition~\ref{consistent-dda-prop}. 

\begin{proposition}\em
Let $\Pi$ be an EDP and $\Sigma$ a set of facts in $\Pi$. 
If $(\Pi\setminus\Sigma)\,\cup\,\overline{\Sigma}$ is consistent, then $\Omega_D$ is not contradictory. 
\begin{proof}\rm
Put $\Pi'=(\Pi\setminus\Sigma)\,\cup\,\overline{\Sigma}$. 
Since facts are set aside in (SD)DA completion, $\Omega_D=\Pi'\cup sdda(\Pi)=\Pi'\cup sdda(\Pi')=SDDA(\Pi')$. 
Suppose that $SDDA(\Pi')$ has the answer set $Lit$. Then $\Pi'$ has the answer set $Lit$ (Proposition~\ref{consistent-dda-prop}). 
Since $\Pi'$ is an EDP, $\Pi'$ is inconsistent, which contradicts the assumption that $\Pi'$ is consistent. 
Hence, the result holds.  
\end{proof}
\end{proposition}

\subsection{Neighborhood Inference} \label{sec:6.3}

{\em Cooperative query answering\/} analyzes the intent of a query and provides associated information relevant to the query (\cite{CCL90}). 
{\em Neighborhood inference\/} (\cite{GGM92}) is a technique used for such a purpose and reasons up/down 
in a taxonomy of atoms to reach neighboring solutions.  
\vspace{.1in}

\begin{example}\rm \label{relax-ex}
A travel agency has flight information represented by the program $\Pi$:  
\begin{eqnarray*}
&& travel(LHR,CDG)\leftarrow flight(AF1681),\\
&& travel(LHR,CDG)\leftarrow flight(BA306),\\
&& travel(NRT,CDG)\leftarrow flight(AF275),\\
&& flight(BA306)\leftarrow,\;\;\;  \neg\,flight(AF1681)\leftarrow,\;\;\; 
flight(AF275)\leftarrow
\end{eqnarray*}
where ``$travel(X,Y)\leftarrow flight(Z)$" means that a flight $Z$ is used for traveling from $X$ to $Y$. 
Suppose that a customer asks the availability of a flight AF1681 from LHR to CDG. 
Unfortunately, no ticket is available on the requested time of a day. 
The agent then proposes an alternative flight BA306 that is still available.  

In this scenario, from the request $flight(AF1681)$ the agent understands that the customer wants to travel from 
London (LHR) to Paris (CDG). 
The request $flight(AF1681)$ is then {\em relaxed\/} to $travel(LHR,CDG)$, and reaches the fact $flight(BA306)$. 
\end{example}

Neighborhood inference consists of two steps: generalization from the body to the head of one rule, 
and specialization from the head to the body of another rule.  The generalization is the inference of affirming the antecedent, 
while the specialization is the inference of affirming the consequent. Then neighborhood inference is realized by combining
AA and AC.  In what follows, we consider a {\em binary program\/} $\Pi$ that consists of ground rules of the form 
``$L_1\leftarrow L_2$" where $L_1$ and $L_2$ are positive/negative literals ($L_2$ is possibly empty). 
$\Pi$ is partitioned into the set of non-factual rules $\Pi_R$ and the set of facts $\Pi_F$, i.e., $\Pi=\Pi_R\cup\Pi_F$. 
\vspace{.1in}

\begin{definition}[neighborhood solution]\rm \label{df-nei-sol}
Let $\Pi$ be a binary program such that $\Pi=\Pi_R\cup \Pi_F$ and $G$ a ground literal representing a request.  
Define 
\begin{eqnarray*}
U&=&\{\,r\,\mid\, r\in \Pi_R \;\mbox{and}\; body^+(r)=\{G\}\,\},\\
V&=&\{\,r'\,\mid\, r'\in\Pi_R\setminus U\;\mbox{and}\; head^+(r')=head^+(r)\; \mbox{for some}\; r\in U\,\}.
\end{eqnarray*} 
If $U$ and $V$ are non-empty and $\Pi_R\cup ac(V)\cup \{G\}$ has a consistent answer set $S$, 
then $S\cap \Pi_F$ is called a {\em neighborhood solution\/} of $G$. 
\end{definition}

By definition, the AC completion is applied to a part of the program, which realizes neighborhood inference based on a request. 

\begin{example}[cont.\ Example~\ref{relax-ex}]\rm 
Given the request $G=flight(AF1681)$,  
$U=\{\, travel(LHR,CDG)\leftarrow flight(AF1681) \,\}$ and 
$V=\{\, travel(LHR,CDG)\leftarrow flight(BA306) \,\}$. 
Then $ac(V)=\{\,flight(BA306)\leftarrow travel(LHR,CDG)\,\}$, and $\Pi_R\cup ac(V)$ becomes 
\begin{eqnarray*}
&& travel(LHR,CDG)\leftarrow flight(AF1681),\\
&& travel(LHR,CDG)\leftarrow flight(BA306),\\
&& travel(NRT,CDG)\leftarrow flight(AF275),\\
&& flight(BA306)  \leftarrow travel(LHR,CDG). 
\end{eqnarray*}
$\Pi_R\cup ac(V)\cup \{G\}$ has the answer set 
\[S=\{\, flight(AF1681),\, flight(BA306),\, travel(LHR,CDG)\,\}.\]  
Hence, $S\cap \Pi_F=\{  flight(BA306) \}$ is a neighborhood solution of $G$. 
\end{example}

\begin{proposition}\em
Let $\Pi$ be a binary program and $G$ a ground literal representing a request. 
If\, $U$ and $V$ in Def.~\ref{df-nei-sol} are non-empty and $AC(\Pi)\cup\{G\}$ is consistent, $G$ has a neighborhood solution. 
\begin{proof}\rm
Since $\Pi_R\cup ac(V)\subseteq AC(\Pi)$, $\Pi_R\cup ac(V)\cup \{G\}$ has a consistent answer set $S$. 
In this case, $G$ has a neighborhood solution $S\cap \Pi_F$. 
\end{proof}
\end{proposition}

\section{Related Work} \label{sec:7}

There is a number of studies on human conditional reasoning in psychology and cognitive science. 
In this section, we focus on related work based on logic programming and its application to common sense reasoning. 

\subsection{Completion} \label{sec:7.1}

The idea of interpreting if-then rules in logic programs as bi-conditional dates back to \cite{Cla78}. 
He introduces {\em predicate completion\/} in normal logic programs (NLPs), which introduces the only-if part of each rule to a program. 
Given a propositional NLP $\Pi$,  Clark completion $Comp(\Pi)$ is obtained by two steps: 
(i) all rules ``$p\leftarrow B_1$", $\ldots$ , ``$p\leftarrow B_k$" in $\Pi$ having the same head $p$ are replaced by 
``$p\leftrightarrow B_1\vee\cdots\vee B_k$", where $B_i$ $(1\le i\le k)$ is a conjunction of literals; and 
(ii) for any atom $p$ appearing in the head of no rule in $\Pi$, add ``$p\leftrightarrow \mathit{false}$". 
The AC completion introduced in this paper extends the technique to the class of GEDP, while 
the result is generally different from Clark completion in NLPs. For instance, given the program:
\[ \Pi_1=\{\, p\leftarrow q,\;\;\; p\leftarrow \,\}, \] 
Clark completion becomes 
\[ Comp(\Pi_1)=\{\, p\leftrightarrow q\vee\top,\;\;\; q\leftrightarrow \bot \,\} \]
where $\top$ and $\bot$ represent {\em true} and {\em false}, respectively. 
$Comp(\Pi_1)$ has the single completion model called a {\em supported model\/} (\cite{ABW88}) $\{p\}$.  
In contrast, \[AC(\Pi_1)=\Pi_1\cup \{\,q\leftarrow p \,\}\] has the answer set $\{p,q\}$. 
The difference comes from the fact that in $Comp(\Pi_1)$, $q$ is identified with {\em false\/} but this is not the case in $AC(\Pi_1)$. 
In Clark completion undefined atoms (i.e., atoms appearing in the head of no rule) are interpreted {\em false}. 
We do not use this type of completion 
because it disturbs the basic type of AC reasoning that infers $q$ from $p$ and ``$p\leftarrow q$".  
Clark completion is extended to normal disjunctive programs by several researchers (\cite{LMR88,AD16,NO18}). 
Those extensions reduce to Clark completion in NLPs, so that they are different from the AC completion.  
We also introduce the DC completion and the DA completion.  
When $\Pi_2=\{\, p\leftarrow not\,q\,\}$, $Comp(\Pi_2)=\{\, p\leftrightarrow \neg q,\;\; q\leftrightarrow\bot \,\}$ has the supported model $\{p\}$. 
On the other hand, \[WDC(\Pi_2)=\Pi_2\cup \{\, q\leftarrow not\,p\,\}\] has two answer sets $\{p\}$ and $\{q\}$. 
When $\Pi_3=\{\, p\leftarrow not\,q,\;\; p\leftarrow q,\;\; q\leftarrow p\,\}$, 
$Comp(\Pi_3)=\{\, p\leftrightarrow q\vee\neg q,\;\; q\leftrightarrow p\,\}$ has the supported model $\{p,q\}$. 
In contrast, 
\[ WDA(\Pi_3)=\Pi_3\cup\{\,not\,p\leftarrow q,not\,q,\;\; not\,q\leftarrow not\,p\,\}\] has no answer set. 
As such, completion semantics introduced in this paper is generally different from Clark completion in NLPs. 
 
The {\em weak completion\/} (\cite{HK09}) leaves undefined atoms {\em unknown\/} under 3-valued logic. 
In the program $\Pi_1=\{\, p\leftarrow q,\;\; p\leftarrow \,\}$, the weak completion becomes 
\[ wcomp(\Pi_1)=\{\, p\leftrightarrow q\vee\top \,\} \]
which is semantically equivalent to $\{\, p\leftrightarrow \top \,\}$. 
Then $p$ is true but $q$ is unknown in $wcomp(\Pi_1)$, which is again different from $AC(\Pi_1)$ that has the answer set $\{p,q\}$. 
In the program $\Pi_2=\{\, p\leftarrow not\,q\,\}$, the weak completion becomes 
\[ wcomp(\Pi_2)=\{\, p\leftrightarrow \neg q \,\} \]
then both $p$ and $q$ are unknown.  In contrast, $WDC(\Pi_2)$ has two answer sets $\{p\}$ and $\{q\}$, 
and $WDA(\Pi_2)$ has the single answer set $\{p\}$. 
 
\subsection{Human Conditional Reasoning} \label{sec:7.2}

\cite{SvL08} formulate human conditional reasoning using Clark's program completion under the three-valued logic of (\cite{Fit85}). 
They represent a conditional sentence ``if $p$ then $q$" as a logic programming rule:%
\footnote{They write it ``$p\wedge \neg ab\rightarrow q$" but we use the standard writing in LP.}
\[ q\leftarrow p\wedge \neg ab\] where $ab$ represents an abnormal atom.  
In this setting, DA is represented as 
\[ \Pi_1=\{\,p\leftarrow\bot,\;\;\; q\leftarrow p\wedge\neg ab,\;\;\; ab\leftarrow\bot\,\}.\]
The rule ``$A\leftarrow\bot$" means that $A$ is a proposition to which the {\em closed world assumption\/} (\cite{Rei78}) is applied. 
If a program does not contain ``$A\leftarrow\bot$", nor any other rule in which $A$ occurs in its head, then 
$A$ is interpreted {\em unknown}. 
Then its completion 
\[  Comp(\Pi_1)=\{\,p\leftrightarrow \bot,\;\;\; q\leftrightarrow p\wedge\neg ab,\;\;\; ab\leftrightarrow \bot\,\}\] derives ``$q\leftrightarrow\bot$". 
On the other hand, completion does not realize AC or DC inference by itself.  In their framework, 
AC is represented as 
\[ \Pi_2=\{\,q\leftarrow\top,\;\; q\leftarrow p\wedge\neg ab,\;\; ab\leftarrow\bot\,\}\] while its completion 
\[ Comp(\Pi_2)=\{\,q\leftrightarrow \top\vee (p\wedge\neg ab),\;\; ab\leftrightarrow\bot\,\}\]
does not derive $p$.  
Likewise, DC is represented as 
\[ \Pi_3=\{\,q\leftarrow\bot,\;\; q\leftarrow p\wedge\neg ab,\;\; ab\leftarrow\bot\,\}\] while its completion 
\[ Comp(\Pi_3)=\{\,q\leftrightarrow \bot\vee (p\wedge\neg ab),\;\; ab\leftrightarrow\bot\,\}\]
does not derive ``$p\leftrightarrow\bot$".  
They then interpret ``$q\leftarrow p\wedge\neg ab$" as an {\em integrity constraint\/} meaning that 
``if $q$ succeeds (resp.\ fails) then `$p\wedge\neg ab$' succeeds (resp.\ fails)" 
to get the AC consequence $p$ (resp.\ DC consequence $\neg p$). 

\cite{SvL08} characterize the suppression task in their formulation. 
The sentence ``If she has an essay to write then she will study late in the library" 
is represented as: 
\[  library\;\leftarrow\; essay\wedge \neg ab_1.\]
Given the negation of antecedent $\neg essay$ (or equivalently, the CWA rule ``$essay\leftarrow\bot$"), the completed program: 
\begin{eqnarray*}
 && library\;\leftrightarrow essay\wedge \neg ab_1,\\
 && ab_1\leftrightarrow\bot,\\
 && essay\leftrightarrow\bot
\end{eqnarray*}
derives ``$library\leftrightarrow\bot$". 
Next, suppose that the conditional with an alternative antecedent: 
``If she has some textbooks to read then she will study late in the library" is given.   The program becomes 
\begin{eqnarray*}
&& library\;\leftarrow\; essay\wedge \neg ab_1,\\
&& library\;\leftarrow\; text\wedge \neg ab_2. 
\end{eqnarray*}
Given the fact $\neg essay$ (or equivalently, the CWA rule ``$essay\leftarrow\bot$"), the completed program: 
\begin{eqnarray*}
 && library\;\leftrightarrow\; (essay\wedge \neg ab_1)\vee (text\wedge \neg ab_2),\\
 && ab_1\leftrightarrow\bot,\\
  && ab_2\leftrightarrow\bot,\\
 && essay\leftrightarrow\bot
\end{eqnarray*}
does not derive ``$library\leftrightarrow\bot$".  Thus, the DA inference is suppressed. 
They also characterize Byrne's suppression of valid inference. 
Suppose the conditional with an additional antecedent: ``If the library stays open then she will study late in the library". The program becomes 
\begin{eqnarray*}
&& library\;\leftarrow\; essay\wedge \neg ab_1,\\
&& library\;\leftarrow\; open\wedge \neg ab_3. 
\end{eqnarray*}
They also introduce interaction of abnormality atoms as
\begin{eqnarray*}
&& ab_1\leftarrow \neg open,\\
&& ab_3\leftarrow \neg essay. 
\end{eqnarray*}
Completing these four rules with ``$ab_1\leftarrow \bot$" and ``$ab_3\leftarrow\bot$" produces 
\begin{eqnarray*}
&& library\leftrightarrow (essay\wedge \neg ab_1)\vee (open\wedge \neg ab_3),  \\
&& ab_1\leftrightarrow \neg open\vee\bot,\\
&& ab_3\leftrightarrow \neg essay\vee\bot,
\end{eqnarray*}
which reduces to 
\[ library\leftrightarrow open\wedge essay.  \]
Then, $essay$ alone does not deduce $library$, so the AA inference is suppressed. 
\cite{SvL08} argue that most people represent the effect of an additional premise formally as ``$p\leftarrow q\wedge r$"  
and that of an alternative premise formally as ``$p\leftarrow q\vee r$". 
This argument coincides with our view addressed in Section~\ref{sec:5.1}. 

\cite{DHR12} point out a technical flaw in the formulation by (\cite{SvL08}). 
In the above example, {\em open\/} and {\em library\/} are unknown ({\sf U}) under the 3-valued logic, 
then the rule ``$library\,\leftarrow\, open\wedge \neg ab_3$" becomes ``${\sf U}\leftarrow {\sf U}$". 
Under the Fitting semantics, however, the truth value of the rule ``${\sf U}\leftarrow {\sf U}$" is ${\sf U}$, 
then it does not represent the truth of the rule ``$library\,\leftarrow\, open\wedge \neg ab_3$". 
To remedy the problem, they employ {\L}ukasiewicz's 3-valued logic which maps ``${\sf U}\leftarrow {\sf U}$" to $\top$. 
\cite{DHR12} also characterize the suppression effects in AC or DC using 
an abductive logic program $\pf{\Pi, \Gamma}$ with abducibles $\Gamma=\{\, p\leftarrow\bot,\;\; p\leftarrow\top\,\}$. 
Consider $\pf{\Pi_1, \Gamma_1}$ where 
\begin{eqnarray*}
\Pi_1:&& library\leftarrow essay\wedge \neg ab_1,\\
&& ab_1\leftarrow\bot,\\
\Gamma_1:&& essay\leftarrow\bot,\;\;\; essay\leftarrow\top,
\end{eqnarray*}
the weakly completed program of $\Pi_1$ becomes  
\begin{eqnarray*}
 && library\leftrightarrow essay\wedge \neg ab_1,\\
 && ab_1\leftrightarrow\bot. 
\end{eqnarray*}
The observation $O=(library\leftrightarrow\top)$ derives 
``$essay\leftrightarrow\top$", then ``$essay\leftarrow\top$" is the skeptical explanation of $O$. 
When the additional rules and abducibles
\begin{eqnarray*}
\Pi_2: && library\leftarrow text\wedge \neg ab_2,\\
&& ab_2\leftarrow\bot,\\
\Gamma_2: && text\leftarrow\bot,\;\;\; text\leftarrow\top
\end{eqnarray*}
are given, the weakly completed program of $\Pi_1\cup\Pi_2$ becomes 
\begin{eqnarray*}
 && library\leftrightarrow (essay\wedge \neg ab_1)\vee (text\wedge\neg ab_2),\\
 && ab_1\leftrightarrow\bot,\\
 && ab_2\leftrightarrow\bot. 
\end{eqnarray*}
The observation $O=(library\leftrightarrow\top)$ derives ``$essay\vee text\leftrightarrow\top$", and there are 
two credulous explanations ``$essay\leftarrow\top$" and ``$text\leftarrow\top$" in $\Gamma_1\cup\Gamma_2$. 
In this case, ``$essay\leftarrow\top$" is not concluded under skeptical reasoning, which represents the suppression of AC. 

Comparing the above mentioned two studies with our approach, there are several differences. 
First, they translate a conditional sentence ``if $p$ then $q$" into the rule ``$q\leftarrow p\wedge \neg\,ab$". 
However, it is unlikely that people who commit logical fallacies, especially younger children (\cite{RCB83}),  
translate the conditional sentence into the rule of the above complex form in their mind. 
We represent the conditional sentence directly as ``$q\leftarrow p$", and assume that people would 
interpret it as bi-conditional depending on the context in which it is used.  
Second, in order to characterize AC or DC reasoning, \cite{SvL08} interpret a conditional sentence as an 
integrity constraint, while (\cite{DHR12}) uses abductive logic programs. 
Our framework does not need a specific interpretation of rules (such as integrity constraints) nor need an extra mechanism 
of abductive logic programs. 
Third, they use a single (weak) completion for all AC/DA/DC reasoning, while we introduce different types of 
completions for each inference.  By separating respective completions, individual inferences 
are realized in a modular way and freely combined depending on their application context. 
For instance, we use AC and (S)DA completion to characterize the suppression task (Section~\ref{sec:5.1}), 
while use AC and (W)DC completion to characterize Wason selection task (Section~\ref{sec:5.2}). 
Fourth, they handle normal logic programs, while our framework can handle a more general class of logic programs 
containing disjunction and two different types of negation.  For instance, consider the rule: 
``$buy\_car \,;\, buy\_house \leftarrow win\_lottery$" (if she wins the lottery, she buys a car or a house). 
When it is known that $buy\_car$ is true, one may infer $win\_lottery$ by AC inference. 
The AC completion realizes such an inference by introducing rules: ``$win\_lottery\leftarrow buy\_car$" and 
``$win\_lottery\leftarrow buy\_house$". 

\cite{CHR21} represent conditionals as in (\cite{SvL08}) and use the weak completion and abductive logic programs as in (\cite{DHR12}). 
They formulate different types of conditionals based on their {\em contexts\/} and 
argue in which case AC or DC is more likely to happen.  More precisely, 
a conditional sentence whose consequent appears to be obligatory given the antecedent is called an {\em obligation conditional}. 
An example of an obligation conditional is that ``{\em if Paul rides a motorbike, then he must wear a helmet}". 
If the consequence of a conditional is not obligatory, then it is called a {\em factual conditional}. 
The antecedent $A$ of a conditional sentence is said to be {\em necessary\/} iff its consequent $C$ cannot be true unless $A$ is true. 
For example, the library's being open is a necessary antecedent for studying in the library. 
Cramer $et\,al.$ argue that AA and DA occur independently of the type of a conditional. On the other hand, 
in AC most people will conclude $A$ from ``$A\Rightarrow C$" and $C$, while the number of people who 
conclude nothing will increase if $A$ is a non-necessary antecedent.  
In DC, most people will conclude $\neg\,A$ from ``$A\Rightarrow C$" and $\neg\,C$, while the number of people who 
conclude nothing will increase if the conditional is factual.  
Those assumptions are verified by questioning participants who do not receive any education in logic beyond high school training. 
They then formulate the situation by introducing the abducible ``$C\leftarrow\top$" if the antecedent is non-necessary, 
and ``$ab\leftarrow\top$" if the conditional is factual.  
In the former case, the observation $C$ does not imply $A$ because the additional ``$C\leftarrow\top$" can make $C$ explainable by itself. 
As a result, $A$ is not a skeptical explanation of $C$. 
In the latter case, the observation $\neg\, C$ does not imply $\neg\, A$ because if one employs the explanation ``$ab\leftarrow\top$" 
then ``$C\leftarrow A\wedge \neg ab$" does not produce ``$C\leftrightarrow A$". 

\cite{DFH22} use logic programming rules to represent different types of conditionals. 
For instance, 
\[  {\sf concl}\leftarrow {\sf prem(x)},\, {\sf sufficient(x)} \]
represents MP that ${\sf concl}$ follows if a sufficient premise is asserted to be true.  By contrast, 
\[  {\sf not\_concl}\leftarrow {\sf not\_prem(x)},\, {\sf necessary(x)} \]
represents DA that ${\sf concl}$ does not follow if a necessary premise is asserted to be false. 
With these rules, Byrne's suppression effect is represented as follows. 
First, given the fact ${\sf prem(essay)}$ and ${\sf sufficient(essay)}$, 
\[ {\sf library}\leftarrow {\sf prem(essay)},\, {\sf sufficient(essay)} \]
implies ${\sf library}$.   Next, given the additional fact ${\sf necessary(open)}$ and 
in the absence of ${\sf prem(open)}$, 
\[ {\sf not\_library}\leftarrow {\sf not\_prem(open)},\, {\sf necessary(open)}\]
has the effect of withdrawing ${\sf library}$. 
In the current study, we do not distinguish different types of conditionals as in (\cite{CHR21}; \cite{DFH22}). 
However, completion is done for individual rules, so we 
could realize {\em partial\/} completion by selecting rules $\Pi'\subseteq \Pi$ that are subject to be completed in practice. 
More precisely, if a program $\Pi$ consists of rules $R_1$ having necessary antecedents and $R_2$ having non-necessary antecedents,  apply AC completion to $R_1$ while keep $R_2$ as they are. The resulting program then realizes AC inference using $R_1$ only. 
Likewise, if a program $\Pi$ consists of rules $R_3$ having obligatory consequents and $R_4$ having factual consequents, 
apply DC completion to $R_3$ while keep $R_4$ as they are. The resulting program then realizes DC inference using $R_3$ only. 
Such a partial completion is also effectively used for commonsense reasoning in this paper (Proposition~\ref{ac-abd-prop}(ii), Definition~\ref{df-nei-sol}).

\subsection{Commonsense Reasoning} \label{sec:7.3}

\cite{CTD91} and (\cite{FK97}) compute abduction by deduction using Clark completion. 
Abduction using AC and DC completion is close to those approaches, while the approach based on Clark completion is 
restricted to normal logic programs (NLPs).  We argued that a (positive) observation $O=\neg G$ is distinguished from 
a negative observation $O=not\,G$, but such a distinction is not considered in NLPs handling only default negation. 
\cite{IS99} introduce {\em transaction programs\/} to compute {\em extended abduction}. 
Extended abduction computes explanations for not only (positive) observations but also negative ones. 
A transaction program is constructed based on the converse of conditionals, and its semantics is operationally given as a fixpoint of the program.  A transaction program is a meta-level specification of the procedure and is different from the current approach. 
Moreover, a transaction program is defined for NLPs only. 

Pereira $et\,al.$ (\citeyear{PAA91,PS17}) realize counterfactual reasoning in logic programming. 
In (\cite{PAA91}), a counterfactual conditional ``$\phi >\psi$" (meaning $\phi\,\Box\!\!\!\rightarrow\,\psi$) is evaluated in a program 
$\Pi$ by adding $\phi$ to $\Pi$ and computing the maximal non-contradictory submodels of the new program. 
Then the counterfactual conditional is true iff $\psi$ holds in all such submodels. 
\cite{PS17} first use abduction to compute possible causes of an event. 
Next, they assume counterfactual assumption and verify whether an expected outcome will happen under possible causes. 
These studies consider extended/normal logic programs under the well-founded semantics and realize counterfactual reasoning 
via program revision or abductive reasoning. Unlike our approach, they do not introduce new rules for DA inference in counterfactual reasoning. 
 
\cite{GGM92} introduce neighborhood inference for query answering in Horn logic programs. 
The scope of a query is expanded by relaxing the specification, which allows a program to return answers related to the original query. 
They introduce a meta-interpreter to realize it and argue for control strategies. 
We show that similar reasoning is simulated in ASP using the AC completion. 

\section{Conclusion}

This paper studies a method of realizing human conditional reasoning in ASP. 
Different types of completions are introduced to realize logically invalid inferences AC and DA as well as a logically valid inference DC. 
They are applied to representing human reasoning tasks in the literature, and are also used for computing common sense reasoning in AI. 
In psychology and cognitive science, empirical studies show that people perform AC, DA or DC inference depending on the 
context in which a conditional sentence is used. 
We could import the results of those studies and encode knowledge in a way that people are likely to use it.  
The proposed theory is used for such a purpose to realize pragmatic inference in ASP and produce results that are close to human reasoning in practice. 

Completions introduced in this paper are defined in a modular way, 
so one can apply respective completion to specific rules of a program according to their contexts. 
They are combined freely and can be mixed in the same program.  
Those completions are general in the sense that they are applied to logic programs containing disjunction, explicit and default negation.  
Since a completed program is still in the class of GEDPs and a GEDP is transformed to a semantically equivalent EDP (\cite{IS98}),  
answer sets of completed programs are computed using existing answer set solvers.


\begin{thebibliography}{}

\bibitem[\protect\citename{Alviano and Dodaro, }2016]{AD16}
Alviano,~M. \& Dodaro,~C. 2016. 
\newblock Completion of disjunctive logic programs. 
\newblock In: {\em Proceedings of the 25th International Joint Conference on Artificial Intelligence}, pp.\ 886--892. 

\bibitem[\protect\citename{Apt {\it et al.}, }1988]{ABW88}
Apt,~K.~R., Blair,~H.~A. \& Walker,~A. 1988. 
\newblock Towards a theory of declarative knowledge. 
\newblock In: J.\ Minker (ed.), {\it Foundations of Deductive Databases and Logic Programming}, 
Morgan Kaufmann, pp.\,89--148. 

\bibitem[\protect\citename{Braine, }1978]{Bra78}
Braine,~M.~D.~S. 1978. 
\newblock On the relation between the natural logic of reasoning and standard logic. 
\newblock {\em Psychological Review} 85:1--21. 

\bibitem[\protect\citename{Braine and O'Brien, }1998]{BO98}
Braine,~M.~D.~S. \& O'Brien,~D.~P.  (eds.) 1998. 
\newblock {\em Mental Logic}. Mahwah, NJ: Erlbaum. 

\bibitem[\protect\citename{Byrne, }{1989}]{Byr89}
Byrne,~R.~M.~J. 1989. 
\newblock Suppressing valid inferences with conditionals. 
\newblock {\em Cognition} 31(1), pp.\ 61--83. 

\bibitem[\protect\citename{Byrne, }{2005}]{Byr05}
Byrne,~R.~M.~J. 2005. 
\newblock {\em The Rational Imagination: How People Create Alternatives to Reality}.  
\newblock Cambridge, MA: MIT Press. 

\bibitem[\protect\citename{Cheng and Holyoak, }1985]{CH85}
Cheng,~P.~W. \& Holyoak,~H.~J. 1985. 
\newblock Pragmatic reasoning schemas. 
\newblock {\em Cognitive Psychology} 17:391--416. 

\bibitem[\protect\citename{Chu {\it et al.}, }1990]{CCL90}
Chu,~W.~W., Chen,~Q. \& Lee,~R.-C.  1990. 
\newblock Cooperative query answering via type abstraction hierarchy. 
\newblock In: S.~M.~Deen (ed.), {\em Cooperating Knowledge Based Systems}, Springer, pp.\ 271--290. 

\bibitem[\protect\citename{Clark, }1978]{Cla78}
Clark,~K.~L. 1978. 
\newblock Negation as failure. 
\newblock In: H.\ Gallaire and J.\ Minker (eds.), {\it Logic and Data Bases}, Plenum Press, pp.\,293--322.  

\bibitem[\protect\citename{Console  {\it et al.}, }1991]{CTD91}
Console,~L.,  Dupr\'{e},~D.~T.  \& Torasso,~P. 1991. 
\newblock On the relationship between abduction and deduction. 
\newblock {\em Journal of Logic and Computation} 1, pp.\ 661--690.

\bibitem[\protect\citename{Cosmides and Tooby, }1992]{CT92}
Cosmides,~L. \& Tooby,~J. 1992. 
\newblock Cognitive adaptions for social exchange. 
\newblock In Barkow, J., Cosmides, L., Tooby, J. (eds.). 
{\it The adapted mind: Evolutionary psychology and the generation of culture}. 
New York: Oxford University Press. pp.\ 163--228.

\bibitem[\protect\citename{Cramer {\it et al.},  }2021]{CHR21}
Cramer,~M., H\"{o}lldobler,~S. \& Ragnl,~M. 2021. 
\newblock Modeling human reasoning about conditionals. 
\newblock In: {\it Proceedings of the 19th International Workshop on Non-Monotonic Reasoning (NMR-21)}, pp.\ 223--232. 

\bibitem[\protect\citename{Dietz {\it et al.}, }2012]{DHR12}
Dietz,~E., H\"{o}lldobler,~S. \& Ragni,~M. 2012. 
\newblock A computational approach to the suppression task. 
\newblock In: {\em Proceedings of the 34th Annual Conference of the Cognitive Science Society}, pp.\ 1500--1505.

\bibitem[\protect\citename{Dietz {\it et al.}, }2022]{DFH22}
Dietz,~E., Fichte,~J.~K. \& Hamiti,~F. 2022. 
\newblock A quantitative symbolic approach to individual human reasoning. 
\newblock In: {\em Proceedings of the 44th Annual Conference of the Cognitive Science Society}, pp.\ 2838--2846.

\bibitem[\protect\citename{Eichhorn {\it et al.}, }2018]{EKR18}
Eichhorn,~C., Kern-Isberner,~G. \& Ragni,~M. 2018. 
\newblock Rational inference patterns based on conditional logic. 
\newblock In: {\it Proceedings of the 32nd AAAI Conference on Artificial Intelligence (AAAI-18)}, pp.\ 1827--1834. 

\bibitem[\protect\citename{Fitting, }1985]{Fit85}
Fitting,~M. 1985. 
\newblock A Kripke-Kleene semantics for logic programs.
\newblock {\em Journal of Logic Programming} 2, pp.\ 295--312. 

\bibitem[\protect\citename{Fung and Kowalski, }1997]{FK97}
Fung,~T.~H. \& Kowalski,~R. 1997. 
\newblock The iff procedure for abductive logic programming. 
\newblock {\em Journal of Logic Programming} 33, pp.\ 151--165.

\bibitem[\protect\citename{Gaasterland  {\it et al.}, }1992]{GGM92}
Gaasterland,~T., Godfrey,~P. \& Minker,~P. 1992. 
\newblock Relaxation as a platform for cooperative answering. 
\newblock {\em Journal of Intelligence Information Systems\/} 1(3/4):293--321.

\bibitem[\protect\citename{Geis and Zwicky, }1971]{GZ71}
Geis,~M.~L. \& Zwicky,~A. 1971. 
\newblock On invited inferences. 
\newblock {\em Linguistic Inquiry\/} 2, pp.\ 561--566. 

\bibitem[\protect\citename{Gelfond and Lifschitz, }1991]{GL91}
Gelfond,~M. \& Lifschitz,~V.  1991. 
\newblock Classical negation in logic programs and disjunctive databases. 
\newblock {\em New Generation Computing\/} 9 (3\&4), pp.\ 365--385. 

\bibitem[\protect\citename{Griggs and Cox, }1982]{GC82}
Griggs,~R.~A. \& Cox,~J.~R. 1982. 
\newblock The elusive thematic-materials effect in Wason's selection task. 
\newblock {\em British journal of psychology\/} 73(3), pp.\ 407--420.

\bibitem[\protect\citename{H\"{o}lldobler and Kencana Ramli, }2009]{HK09}
H\"{o}lldobler,~S. \& Kencana Ramli,~C.~D. 2009. 
\newblock Logic programs under three-valued Lukasiewicz's semantics. 
\newblock In: {\em Proceedings of the 25th International Conference on Logic Programming, 
Lecture Notes in Computer Science}, vol.\ 5649, Springer, pp.\ 464--478. 

\bibitem[\protect\citename{Horn, }2000]{Horn00}
Horn,~L.~R. 2000. 
\newblock From if to iff: conditional perfection as pragmatic strengthening. 
\newblock {\em Journal of Pragmatics} 32, pp.\ 289--326. 

\bibitem[\protect\citename{Inoue and Sakama, }1998]{IS98}
Inoue,~K. \& Sakama,~C. 1998. 
\newblock Negation as failure in the head. 
\newblock {\em Journal of Logic Programming\/} 35(1), pp.\ 39--78. 

\bibitem[\protect\citename{Inoue and Sakama, }1999]{IS99}
Inoue,~K. \& Sakama,~C. 1999. 
\newblock Computing extended abduction through transaction programs. 
\newblock {\em Annals of Mathematics and Artificial Intelligence\/} 25(3\&4), pp.\ 339--367. 

\bibitem[\protect\citename{Johnson-Laird, }1983]{JL83}
Johnson-Laird,~P.~N. 1983. 
\newblock {\em Mental models}. 
\newblock Cambridge, MA: Harvard University Press. 

\bibitem[\protect\citename{Kakas {\it et al.}, }1992]{KKT92}
Kakas,~A.~C., Kowalski,~R.~A. \& Toni,~F. 1992. 
\newblock Abductive logic programming. 
\newblock {\em Journal of Logic and Computation\/} 2(6), pp.\ 719--770.
 
\bibitem[\protect\citename{Kowalski, }2011]{Kow11}
Kowalski,~R.~A. 2011. 
\newblock {\em Computational Logic and Human Thinking: How to be Artificially Intelligent}. 
\newblock Cambridge University Press. 

\bibitem[\protect\citename{Lifschitz and Woo, }1992]{LW92}
Lifschitz,~V. \& Woo,~T.~Y.~C. 1992. 
\newblock Answer sets in general nonmonotonic reasoning (preliminary report). 
\newblock In: B.\ Nebel, C.\ Rich, and W.\ Swartout (eds.), 
{\it Principles of Knowledge Representation and Reasoning: 
Proceedings of the Third International Conference}, Morgan Kaufmann, pp.\,603--614.  

\bibitem[\protect\citename{Lifschitz {\it et al.}, }2001]{LPV01}
Lifschitz,~V., Pearce,~D. \& Valverde,~A.  2001. 
\newblock Strongly equivalent logic programs.
\newblock {\em ACM Transactions on Computational Logic\/} 2,  pp.\ 526--541. 

\bibitem[\protect\citename{Lewis, }1973]{Lew73}
Lewis,~D. 1973. 
\newblock {\em Counterfactuals}. Blackwell Publishing.

\bibitem[\protect\citename{Lobo {\it et al.}, }1988]{LMR88}
Lobo, J., Minker, J. \& Rajasekar, A. 1988. 
\newblock Weak completion theory for non-Horn programs. 
\newblock In: R. A. Kowalski and K. A. Bowen (eds.), 
{\em Proceedings of the Fifth lnternational Conference and Symposium on Logic Programming}, 
MIT Press, Cambridge, MA, pp.\ 828--842. 

\bibitem[\protect\citename{Nieves and Osorio, }2018]{NO18}
Nieves,~J.~C. \& Osorio,~M.  2018. 
\newblock Extending well-founded semantics with Clark's completion for disjunctive logic programs. 
\newblock Hindawi Scientific Programming, Article ID 4157030. 

\bibitem[\protect\citename{Oaksford and Chater, }2001]{OC01}
Oaksford,~M. \& Chater,~N.  2001. 
\newblock The probabilistic approach to human reasoning. 
\newblock {\em Trends in Cognitive Science} 5, pp.\ 349--357. 

\bibitem[\protect\citename{Pereira {\it et al.}, }1991]{PAA91}
Pereira,~L.~P., Apar\'{i}cio,~J.~N. \& Alferes,~J.~J.  1991. 
\newblock Counterfactual reasoning based on revising assumptions. 
\newblock In: {\em Logic Programming, Proceedings of the 1991 International Symposium}, MIT Press, pp.\ 566--577. 

\bibitem[\protect\citename{Pereira and Saptawijaya, }2017]{PS17}
Pereira,~L.~P. \& Saptawijaya,~A. 2017. 
\newblock Counterfactuals in logic programming. 
\newblock In: {\em Programming Machine Ethics}, Springer, pp.\ 81--93. 

\bibitem[\protect\citename{Reiter, }1978]{Rei78}
Reiter,~R. 1978. 
\newblock On closed world data bases. 
\newblock In: H.\ Gallaire and J.\ Minker (eds.), {\it Logic and Data Bases}, Plenum Press, pp.\ 119--140. 

\bibitem[\protect\citename{Reiter, }1980]{Rei80}
Reiter,~R. 1980. 
\newblock A logic for default reasoning. 
\newblock {\it Artificial Intelligence\/} 13:81--132.  

\bibitem[\protect\citename{Rumain {\it et al.}, }1983]{RCB83}
Rumain,~B., Connell,~J., \& Braine,~M.~D.~S. 1983. 
\newblock Conversational comprehension processes are
responsible for reasoning fallacies in children as well as adults: IF is not the biconditional.
\newblock {\em Developmental Psychology\/} 19:471--481.

\bibitem[\protect\citename{Sakama and Inoue, }1995]{SI95}
Sakama,~C. \& Inoue,~K. 1995. 
\newblock Paraconsistent stable semantics for extended disjunctive programs. 
\newblock {\em Journal of Logic and Computation\/} 5(3):265--285.

\bibitem[\protect\citename{Stenning and Lambalgen, }2008]{SvL08}
Stenning,~K. \& van Lambalgen, M. 2008. 
\newblock {\em Human Reasoning and Cognitive Science}, MIT Press. 

\bibitem[\protect\citename{Wason, }1968]{Was68}
Wason,,~P.~C. 1968. 
\newblock Reasoning about a rule. 
\newblock {\it Quarterly Journal of Experimental Psychology\/} 20:273--281.  

\bibitem[\protect\citename{Wason and Shapiro, }1971]{WS71}
Wason,~P.~C. \& Shapiro,~D. 1971. 
\newblock Natural and contrived experience in a reasoning problem. 
\newblock {\it Quarterly Journal of Experimental Psychology\/} 23:63--71.

\end{thebibliography}
\end{document}